\documentclass[journal]{IEEEtran}

\usepackage{bm}
\usepackage{amsfonts}
\usepackage{amsmath}
\usepackage{amssymb}
\usepackage{lipsum}

\usepackage{multirow}
\usepackage{booktabs}
\usepackage{array}

\usepackage[noadjust]{cite}

\usepackage{graphicx}
\graphicspath{{./fig/}}

\usepackage{pgfplots}
\pgfplotsset{compat=1.15}
\definecolor{dy}{RGB}{218, 165, 32}

\usepackage[hidelinks]{hyperref}

\usepackage[caption=false]{subfig}
\usepackage{array}
\newcommand{\PreserveBackslash}[1]{\let\temp=\\#1\let\\=\temp}
\newcolumntype{C}[1]{>{\PreserveBackslash\centering}p{#1}}
\newcolumntype{R}[1]{>{\PreserveBackslash\raggedleft}p{#1}}
\newcolumntype{L}[1]{>{\PreserveBackslash\raggedright}p{#1}}

\DeclareMathOperator*{\diag}{diag}

\newcommand{\R}{\mathbb{R}}

\newcommand{\X}{\mathcal{X}}
\newcommand{\Z}{\mathcal{Z}}

\newcommand{\G}{\mathcal{G}}
\newcommand{\V}{\mathcal{V}}
\newcommand{\E}{\mathcal{E}}

\newcommand{\N}{\mathcal{N}}
\newcommand{\U}{\mathcal{U}}

\DeclareMathOperator{\EX}{\mathbb{E}}

\newcommand{\T}{\mathcal{T}}

\newcommand{\tp}[1]{\textcolor{green}{#1}}
\newcommand{\fp}[1]{\textcolor{blue}{#1}}
\newcommand{\fn}[1]{\textcolor{red}{#1}}
\newcommand{\tn}[1]{\textcolor{black}{#1}}

\title{Joint Detection and Localization of Stealth False Data Injection Attacks in Smart Grids\\ using Graph Neural Networks}

\date{\today}

\begin{document}
	\author{
		Osman~Boyaci,
		Mohammad~Rasoul~Narimani,
		Katherine~Davis,
		Muhammad~Ismail,
		Thomas~J~Overbye, and
		Erchin~Serpedin
		\thanks{
			Manuscript received May 23, 2021; revised August 27, 2021 and September 30; accepted October 2, 2021.
			
			O. Boyaci, K. Davis, T. J. Overbye and E. Serpedin are with the Department of Electrical and Computer Engineering, Texas A\&M University, College Station, TX 77843 USA (email: osman.boyaci@tamu.edu; katedavis@tamu.edu; overbye@tamu.edu; eserpedin@tamu.edu)
		
			M. R. Narimani is with the College of Engineering, Arkansas State University, Jonesboro, AR 72404 USA (e-mail: mnarimani@astate.edu).
			
			M. Ismail is with the Department of Computer Science, Tennessee Tech University, Cookeville, TN 38505 USA (email: mismail@tntech.edu)
		}
		\thanks{This work was supported by NSF under Award Number 1808064.}
	}
	\maketitle
	\begin{abstract}\label{abstract} 
	False data injection attacks (FDIA) are a main category of cyber-attacks threatening the security of power systems.
	Contrary to the detection of these attacks, less attention has been paid to identifying the attacked units of the grid.
	To this end, this work jointly studies detecting and localizing the stealth FDIA in power grids. Exploiting the inherent graph topology of power systems as well as the spatial correlations of measurement data, this paper proposes an approach based on the graph neural network (GNN) to identify the presence and location of the FDIA.
	The proposed approach leverages the auto-regressive moving average (ARMA) type graph filters (GFs) which can better adapt to sharp changes in the spectral domain due to their rational type filter composition compared to the polynomial type GFs such as Chebyshev.
	To the best of our knowledge, this is the first work based on GNN that automatically detects and localizes FDIA in power systems.
	Extensive simulations and visualizations show that the proposed approach outperforms the available methods in both detection and localization of FDIA for different IEEE test systems.
	Thus, the targeted areas can be identified and preventive actions can be taken before the attack impacts the grid.
	\end{abstract}


	\section*{Nomenclature}
	\addcontentsline{toc}{section}{Nomenclature}
	\begin{IEEEdescription}[\IEEEusemathlabelsep\IEEEsetlabelwidth{$12345678$}]
		\item[$\bm{P_i}$ + $j \bm{Q_i}$] Complex power injection at bus $i$.
		\item[$\bm{P_{ij}}$ + $j \bm{Q_{ij}}$] Complex power flow between bus $i$ and $j$.
		\item[$\bm{V_i}, \bm{\theta_i}$] Voltage magnitude and phase angle of bus $i$.
		\item[$n, \ m$] Number of buses, number of measurements.
		\item[$\X \in \R^{n}$] State space.
		\item[$\Z \in \R^{m}$] Measurement space.
		\item[$\bm{x} \in \X$] A state vector.
		\item[$\bm{\hat{x}} \in \X$] Original state vector without an attack.
		\item[$\bm{\check{x}} \in \X$] False data injected state vector.
		\item[$\bm{z} \in \Z$] A measurement vector.
		\item[$\bm{z_o} \in \Z$] Original measurement vector.
		\item[$\bm{z_a} \in \Z$] Attacked measurement vector.
		\item[$\bm{a} \in \Z$] Attack vector.
		\item[$h(\bm{x})$] Nonlinear measurement function at $\bm{x}$.
		\item[$\T$] Attacker's target area to perform FDI attack.
		\item[$\bm{W} \in R^{n \times n} $] Weighted adjacency matrix.
		\item[$\bm{D} \in R^{n \times n} $] $\bm{D}_{ii} = \sum_{j} \bm{W}_{ij}$ Diagonal degree matrix. 		
		\item[$\bm{\Lambda} \in \R^{n \times n}$] $=\diag[\lambda_1, \ldots, \lambda_{n}]$ Graph Fourier frequencies.
		\item[$\bm{U} \in \R^{n \times n}$] $=[\bm{u}_1, \ldots, \bm{u}_{n}]$  Graph Fourier basis.
		\item[$\bm{L} \in \R^{n \times n} $] $ = \bm{U} \bm{\Lambda} \bm{U}^T$ Normalized graph Laplacian.		
	\end{IEEEdescription}

	\section{Introduction}\label{introduction} 
	
	Smart grids integrate Information and Communication Technologies (ICT) into large-scale power networks to generate, transmit, and distribute electricity more efficiently~\cite{yu2016smart}. Remote Terminal Units (RTUs) and Phasor Measurement Units (PMUs) are utilized to acquire the physical measurements and deliver them to the Supervisory Control and Data Acquisition Systems (SCADAs). Then, the ICT network transfers these measurements to the application level where the power system operators process them and take the necessary actions \cite{davis2012power}. As a direct consequence, power system reliability is determined by the accuracy of the steps along this cyber-physical pipeline \cite{sridhar2011cyber}.
	Power system state estimation (PSSE) modules employ these measurements to estimate the current operating point of the grid~\cite{abur2004power} and thus the integrity and trustworthiness of the measurements are crucial  for proper operation of power systems. In addition, the accuracy of power system analysis tools such as energy management, contingency and reliability analysis, load and price forecasting, and economic dispatch depends on these measurements~\cite{giannakis2013monitoring}. Thus, power system operation strongly depends on the accuracy of the measurements and the integrity of their flow through the system. Therefore, metering devices represent highly attractive targets for adversaries that try to obstruct the grid operation by corrupting the measurements.
	
	By disrupting the integrity of measurement data, false data injection attacks (FDIAs) constitute a considerable cyber-physical threat. More specifically, an adversary injects some false data to the measurements in order to mislead the PSSE and force  it to converge to another operating point. Since the state of the power system is miscalculated by using  these false data, any action taken by the grid operator based on the  false operating point can lead to serious physical consequences including systematic problems and failures \cite{liang2016review}.
	In traditional power grids, the largest normalized residual test (LNRT) is employed within the bad data detection (BDD) module along with PSSE to detect the ``bad'' measurement data \cite{abur2004power}. Nevertheless, a designed false data injected measurement can 
	bypass the BDD.
	In particular, \cite{liu2011false,davis2012power} show that by satisfying the power flow equations, an intruder can create an unobservable (stealth) FDIA and bypass the BDD if s/he has sufficient information about the grid.
	Various methods have been proposed to alert the grid operator about the presence of the FDIA without providing any information about the attack location \cite{musleh2019survey, sayghe2020survey}. 
	Localizing the attack is  crucial for power system operators since they can take preventive action such as isolating the under-attack buses and re-dispatching the system accordingly.
	Therefore, this paper focuses jointly on detection and localization of the FDIA in power systems.

	\subsection{Related Works}
	In general, there are two main approaches to detect and localize the FDIAs: model-based and data driven approaches \cite{musleh2019survey}.
	In the model-based methods \cite{nudell2015real, khalaf2018joint, drayer2019cyber, luo2020interval, hasnat2020detection}, a model for the system is built and its parameters are estimated to detect the FDIAs.
	Since there is no training, these methods do not require the historical data.
	However, the detection delays, scalability issues and threshold tuning steps can limit the performance and usability of the model-based approaches \cite{sayghe2020survey}.
	Conversely, the data-driven methods~\cite{jevtic2018physics} are system independent and require historical data and a training procedure.
	However, they provide scalability and real time compatibility due to the excessive training.
	Data driven methods, machine learning (ML) \cite{machine_learning}, in particular, offer superior performances to detect FDIAs in power systems as the historical datasets are growing \cite{musleh2019survey, sayghe2020survey}.
	Therefore, we employ a data driven approach in this work for detecting and localizing FDIAs in power systems.

	While there has been a great deal of research on detection of FDIAs, only a few attempts have been made to localize these attacks \cite{nudell2015real, khalaf2018joint, drayer2019cyber, luo2020interval, hasnat2020detection, jevtic2018physics}.
	Since localization of FDIAs is relatively a newer research subject compared to detection of these attacks, the current approaches proposed in literature suffer from some limitations.
	A multistage localization algorithm based on graph theory results is proposed in \cite{nudell2015real} to localize the attack at cluster level.
	Nevertheless, the low resolution hinders the benefits of localization in cluster level algorithms.
	In \cite{khalaf2018joint}, a model-driven analytical redundancy approach utilizing Kalman filters is presented for joint detection and mitigation of FDIA in AGC systems. 
	In their model, the authors of  \cite{khalaf2018joint} first determine a threshold using the Mahalanobis norm of the residuals of the non-attacked situation.
	Any residue larger than the threshold is regarded as an attacked sample.
	Apart from the manual threshold optimization steps, detection times are at the range of seconds in their estimation based models. 	
	A generalized modulation operator that is applied on the states of the system is presented as an ongoing work in a brief announcement in \cite{drayer2019cyber} to localize the FDIAs in power systems. Yet, the results are not published as of today. 
	Authors in \cite{luo2020interval} present an internal observer-based detection and localization method for FDIA in power systems. They create and assign an interval observer to each measurement device and construct a customized logic localization judgment matrix to detect and localize the FDIA. Nevertheless, their average detection delay is more than 1.1 seconds, which can highly limit their usability in a real life scenario. Lack of scalability and the need for a custom solution requiring manual labor represent additional limitations of this method.
	A Graph Signal Processing (GSP) based approach is developed in \cite{hasnat2020detection} to detect and localize FDIAs using the Graph Fourier Transform (GFT), local smoothness, and vertex-frequency energy distribution methods. Hovewer, the  random and easily detectable attacks employed to test their models do not comprehensively assess the actual performance of the models. Besides, manual threshold tuning of graph filters (GFs) brings extra effort for their proposed methods. 
	Authors in \cite{jevtic2018physics} propose physics- and learning-based approaches to detect and localize the FDIAs in automatic generation control (AGC) of power systems.
	While the physics-based method relies on interaction variables, the learning-based approach exploits the historical Area Control Error (ACE) data, and utilizes a Long Short Term Memory (LSTM) Neural Network (NN) to generate a model for learning the data pattern.
	Nevertheless, \cite{jevtic2018physics}  reports results limited to a 5-bus system and assumes training an LSTM model for each measurement.
	Thus, the limited number of components deeply confines the large scale attributes of the proposed method.
	Furthermore, training a separate detector for each bus extremely increases the overall model complexity for large systems and reduces its suitability for real world applications.
	
	\subsection{Motivation}
\begin{figure}[h!] 
		\centering
		\newcommand{\wdt}{0.47} 
		\subfloat[Processing an image having 30 pixels in Euclidean space. Nodes and and edges represent pixels and their neighbors in the image, respectively. \label{fig:euclid}]{
			\centering
			\includegraphics[width=\wdt\linewidth]{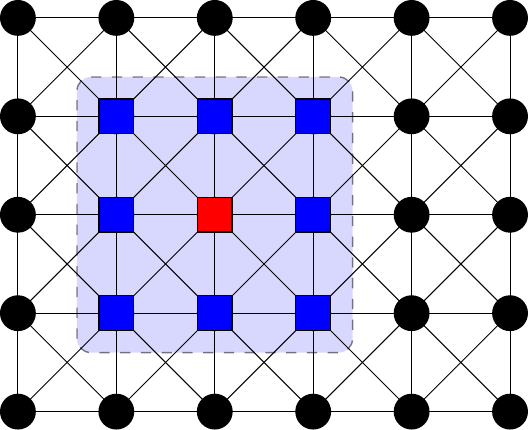}
		}
		\subfloat[Processing a power grid having 30 buses in non-Euclidean space. Nodes and edges represent buses and lines of the grid, respectively. \label{fig:non_euclid}]{
			\centering
			\includegraphics[width=\wdt\linewidth]{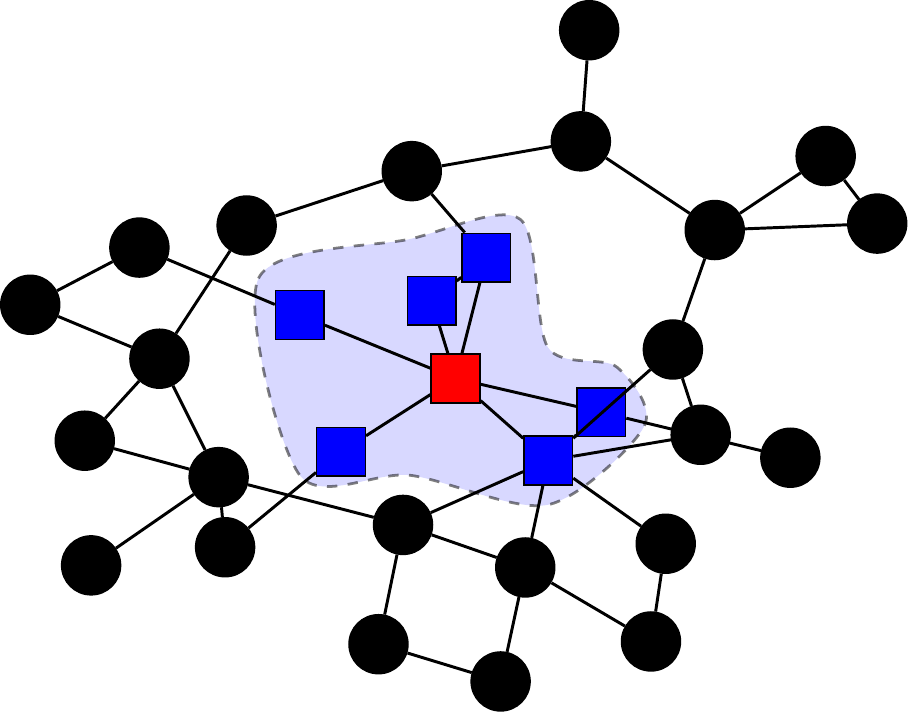}
		}
		\caption{Demonstration of signal processing in Euclidean (\ref{fig:euclid}) and non-Euclidean (\ref{fig:non_euclid}) spaces with an image and a power grid signal \cite{wu2020comprehensive}. Neighbor nodes (blue) of a node (red) are ordered and constant in size for the image having 30 pixels in 2D Euclidean space. In contrast, they are unordered and variable in size for the IEEE 30-bus system in Non-Euclidean space. Therefore, in order to efficiently model the spatial correlation of the power grid, graph type approaches that consider the topology of the underlying systems such as GSP and GNN are necessary.}
		\label{fig:topology}
	\end{figure}

	Due to their graph-based topology, graph structural data such as social networks, traffic networks, and electric grid networks cannot be modeled efficiently in the Euclidean space and require graph-type architectures \cite{wu2020comprehensive}.
	Processing (filtering) an image having 30 pixels and a power grid having 30 buses are demonstrated in Fig.~\ref{fig:topology}.
	Since nodes are ordered and have the same number of neighbors for image data, it can be processed in a 2D Euclidean space.
	For example a sliding kernel can easily capture the spatial correlations of pixels in this Euclidean space.
	Conversely, neighborhood relationships are unordered and vary from node to node in a graph signal \cite{wu2020comprehensive}.
	Therefore, graph signals need to be processed in non-Euclidean spaces determined by the topology of the graph.
	In fact, as a highly complex graph structural data, smart grid signals require graph type architectures such as GSP or GNN to exploit the spatial correlations of the grid.
	
	GSP has emerged in the past few years to deal with the data in non-Euclidean spaces \cite{ortega2018graph}.
	A few researchers designed GFs to detect and localize the FDIA \cite{drayer2019cyber, hasnat2020detection} using GSP.
	However, manual tailoring of the filters and detection thresholds substantially limits the applicability and efficiency of GSP.
	Conversely, GNN, as a data-driven counterpart of the GSP, eliminates the custom design steps and provides an end-to-end design that exploits the spatial locality dictated by the historical data.
	Similar to the classical signal processing, a graph signal is first converted into the spectral domain by GFT, then its Fourier coefficients are multiplied with those filter weights and finally the signal transformed back into the vertex domain by the inverse GFT \cite{shuman2013emerging}.
	To circumvent this spectral decomposition and domain transformation, polynomial GFs are proposed in \cite{defferrard2016convolutional} in which localized filters are learned directly in the vertex domain \cite{bianchi2021graph}.
	For a polynomial GF, the output of each vertex $v$ is only dependent on the $K$-hop neighborhood of $v$ and its spectral response is a $K$-order polynomial.
	Polynomial GFs, which are also referred to as finite impulse response (FIR) GFs due to the local information sharing, perform a weighted moving average (MA) filtering \cite{shi2015infinite, tremblay2018design}.
	However, FIR GFs may require a high degree polynomial to capture the global structure of the graph. In fact, interpolation and extrapolation performance of high degree polynomials are unsatisfactory \cite{shi2015infinite}, and they are not ``flexible'' enough to adapt to sudden changes in the spectral domain \cite{bianchi2021graph}. 
	To overcome this limitation, infinite impulse response (IIR) type GFs performing Auto-Regressive Moving Average (ARMA) are proposed in \cite{shi2015infinite}.
	Contrary to FIR GFs, IIR GFs have rational type spectral responses.
	Therefore, IIR GFs can implement more complex responses with a low degree of polynomials both in the numerator and denominator since rational functions have better performance compared to polynomial ones in terms of interpolation and extrapolation capabilities \cite{shi2015infinite, bianchi2021graph}.

	Detection and localization of FDIA can be a challenging task if an intruder has `enough' information about the grid to create a stealth attack \cite{liu2011false}.
	S/he can hide an attack vector into an honest sample if the topology of the grid is ignored.
	Moreover, s/he can design an attack vector so that a malicious sample can be indistinguishable from an honest one if the spatial correlations of grid data are not well captured or the designed GFs do not satisfy the required spectral response.
	Thus, we design an GNN based model by utilizing ARMA GFs to be able to fit sharp changes in the spectral domain of the grid.
	Filter weights are learned automatically during training by an end-to-end data-driven approach.
	To compare our results with the existing data-driven techniques, we utilized several models to jointly detect and localize the FDIA.
	Moreover, for a fair comparison, the Bayesian hyper-parameter optimization technique is employed to all models for tuning the models' hyperparameters such as number of layers, neurons, etc.
	
	\subsection{Contributions and Paper Organization}	
	The contributions of this work are outlined as follows:
	\begin{itemize}
		
		\item To properly capture the spatial correlations of the smart grid data in a non-Euclidean space, we utilize IIR type ARMA GFs which provide more flexible frequency responses compared to FIR type Chebyshev GFs.
		It is demonstrated on IEEE 118- and 300-bus test systems that ARMA GFs better approximate the desired filter response compared to CHEB GFs for the same filter order by comparing their empirical frequency responses when approximating an ideal band pass filter. 

		\item To precisely test our proposed method, we generate a dataset for each test system with 1-minute intervals using several FDIA generation algorithms in the literature as well as our optimization-based FDIA method developed in our previous paper~\cite{boyaci2021graph}.

		\item To automatically determine the unknown filter weights by an end-to-end data-driven approach, we propose a scalable, ARMA GF-based GNN model that jointly detects and localizes the FDIAs in a few milliseconds.
		The proposed architecture efficiently predicts the presence of the attack for the whole grid and for each bus separately.
		
		\item To fairly compare the proposed method with the currently available approaches, we implement the other data-driven models in the literature and compare our detection and localization results with them.
		Hyperparameters of the models are tuned systematically using the Bayesian hyper-parameter optimization technique.

		\item To adequately assess the localization performance, we evaluate the localization results, using both sample wise and node wise comparisons.
		For instance, although sample wise localization could yield fairly high accuracy for the entire system, the  same set of nodes could be missed or falsely alarmed at each sample. If revealed, these nodes could be easily targeted by the intruders.

		\item To better analyze and visualize the multidimensional data processed by the implemented models, we embed them into a two dimensional (2D) space using the t-SNE algorithm~\cite{tsne}. By visually inspecting the output of models' intermediate layers in 2D, it is verified that the ARMA GNN based model preserves the structure of the data, and hence gives better detection performance.

	\end{itemize}
	
	The rest of this paper is organized as follows. Section~\ref{problem} presents the problem formulation. Section~\ref{methods} proposes the approach for the joint detection and localization of FDIA. Numerical results are presented in Section~\ref{details}. Section~\ref{conclusion} finally concludes the paper. 

	\section{Problem Formulation}\label{problem} 
	The system state $\bm{x}$  ($V_i$ and $\theta_i$ at each bus $i$) 
	is estimated using the PSSE module. The PSSE iteratively solves the optimization problem in~\eqref{eq:psse} phrased as a weighted least squares estimation (WLSE) using the complex power measurements $\bm{z}$ collected in noisy conditions by RTUs and PMUs: 
	\begin{equation} \label{eq:psse}
	\bm{\hat{x}} = \min_{\bm{x}} (\bm{z} - h(\bm{x}))^T \bm{R}^{-1} (\bm{z} - h(\bm{x})),
	\end{equation}	
	where $\bm{R}$ represents measurements' error covariance matrix and $\bm{z}$ includes $\bm{P_i}$, $\bm{Q_i}$, $\bm{P_{ij}}$, and $\bm{Q_{ij}}$.
	
	FDIA aim to deceive the PSSE by deliberately injecting false data $\bm{a}$ into some of the original measurements $\bm{z_o}$ in such a way that the state vector $\bm{x}$ converges to another point in the state space of the system. Formally,
	\begin{equation}
	\bm{z_o} = h(\bm{\hat{x}}), \ \
	\bm{z_a} = \bm{z_o} + \bm{a} = h(\bm{\check{x}})
	\end{equation} 
	which means if an adversary can design $\bm{a} = h(\bm{\check{x}}) - h(\bm{\hat{x}})$, s/he can change the system state from $\bm{\hat{x}}$ to $\bm{\check{x}}$ without being detected by the LNRT based traditional BDD systems.

	In general, an adversary tries to change specific measurement(s) in the power system in order to maximize the damage to the grid and at the same time minimize the probability of being detected. To this end, s/he alters some other measurement(s) connected to the targeted meter(s) since each $\bm{x}$ relates to multiple $\bm{z}$ through $\bm{z} = h(\bm{x})$. In order to reflect this constraint and to be realistic, we assume that an adversary targets a specific area of grid represented by $\T$ and crafts the attack vector $\bm{a}$ by changing the measurements denoted by $\T_z$ to spoil the state variables represented by $\T_x$ in this area.

	\begin{figure}[h!]
		\centering
		\includegraphics[width=0.30\textwidth]{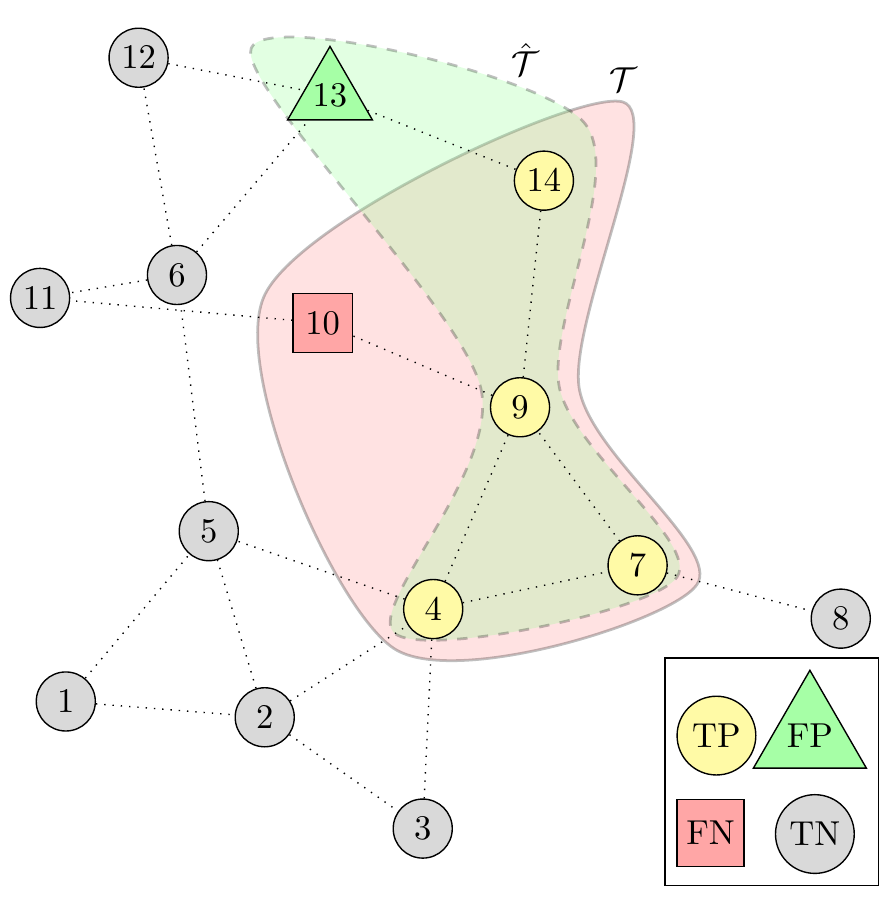}
		\caption{Visualization of an attack and its prediction on the example IEEE 14 bus system where the actual $\T=\{4,7,9,10,14\}$ and predicted $\hat{\T}=\{4,7,9,13,14\}$ areas are enclosed with the solid red and dashed green surfaces, respectively. True positives $\T\cap\hat{\T} = \{4,7,9,14\}$, false positives $\T'\cap\hat{\T} = \{13\}$, false negatives $\T\cap\hat{\T}'=\{10\}$, and true negatives $\T'\cap\hat{\T}'=\{1,2,3,5,8,6,11,12\}$ are represented by yellow circles, green triangles, red squares, and black circles, respectively. In this example, the presence of the attack is correctly predicted. Nevertheless, attack to the bus 10 is missed and bus 13 is falsely alarmed even though  it is not under attack.}
		\label{fig:problem}
	\end{figure}

	The grid operator, in contrast, aims to detect those attacks and localize the attacked buses if there are any. Therefore, we formulate the FDIA detecting and localization problem as a \emph{multi-label} classification task where each bus has a binary label indicating the presence of attack with true label 1. We also reserve an extra binary label for the whole grid to denote the attack presence with true label 1. Fig.~\ref{fig:problem} clarifies the proposed multi-label classification approach by depicting the actual and predicted under attack buses for an exemplary attack on the IEEE 14-bus test system.

	\section{Joint Detection and Localization of FDIA}\label{methods} 
	Connected, undirected and weighted graph $\G = (\V, \E, \bm{W})$ having a finite set of vertices $\V$ with $|\V| = n$, a finite set of edges $\E$, and a weighted adjacency matrix $\bm{W} \in \R^{n \times n}$ can be used to represent the topology of a smart power grid \cite{ortega2018graph}.
	In this representation, buses correspond to vertices $\V$, branches and transformers corresponds to edges $\E$ and line admittances correspond to $\bm{W}$.
	Similarly, a signal or a function $f:\V \rightarrow \R $ in $\G$ is  represented by a vector $\bm{f} \in \R^n$, where the element $i$ of the vector corresponds to a scalar at the vertex $i \in \V$.

	\subsection{Spectral Graph Filters}
	In spectral graph theory, the normalized Laplacian operator
	$ \bm{L} = \bm{I_n} - \bm{D}^{-1/2} \bm{W} \bm{D}^{-1/2} = \bm{U} \bm{\Lambda} \bm{U}^T \in \R^{n \times n} $
	plays an important role for graph $\G$ where $\bm{D}$ and $\bm{I_n} \in \R^{n \times n}$ represent the degree and identity matrices, respectively. The columns $\bm{u}_i \in \R^{n \times 1}$ of matrix $\bm{U} = [\bm{u}_1, \ldots, \bm{u}_{n}] \in \R^{n \times n}$ store the $n$ orthonormal eigenvector $\bm{u_i}$ and constitute the graph Fourier basis.
	Diagonal matrix $\bm{\Lambda} = \diag([\lambda_1, \ldots, \lambda_{n}]) \in \R^{n \times n}$ captures the $n$ eigenvalues representing the graph Fourier frequencies \cite{ortega2018graph}. 
	Analogously to the classical Fourier Transform, Graph Fourier Transform (GFT) transforms a vertex domain signal into the spectral domain: the forward and inverse GFT are defined by $\bm{\tilde{X}} = \bm{U}^T \bm{X}$, and $\bm{X} = \bm{U} \bm{\tilde{X}}$, where $\bm{X}$ and $\bm{\tilde{X}} \in \R^{n \times f}$ denote the vertex and spectral domain signals with $f$ features at each node, respectively \cite{ortega2018graph}. 
	In fact, $\bm{X}$ is filtered by a GF $h$:
	\begin{equation} \label{eq:gsp}
		\bm{Y} = h \ast \bm{X} = h(\bm{L}) \bm{X} = \bm{U} h(\bm{\Lambda}) \bm{U}^T \bm{X}
	\end{equation}
	by first converting the vertex domain signal $\bm{X}$ into the spectral domain using the forward GFT, then scaling the Fourier components by $h(\bm{\Lambda}) = \diag[h(\lambda_1), \ldots, h(\lambda_{n})]$, and finally reverting it back to the vertex domain by the inverse GFT \cite{ortega2018graph}.
	For example, $\bm{X}$, $h$, and $\bm{Y}$ may correspond to bus injections values with high frequency noise, a low pass GF and filtered bus injections values, respectively in eq. (3). 
	Nonetheless, this spectral filtering is not spatially localized since each $\lambda_i$ is processed for each node.
	Besides its computational complexity is high due to eigenvalue decomposition (EVD) of $\bm{L}$ and the matrix multiplications with $\bm{U}$ and $\bm{U}^T$.
	
	\subsection{Polynomial Graph Filters}
	To localize spectral filters and reduce their complexity, polynomial spatial filters $h_{POLY}(\lambda) = \sum_{k=0}^{K-1} a_k \lambda^k$ are proposed to approximate the required filter response \cite{defferrard2016convolutional}.
	Since only $K$-hop neighbors of $v$ are considered to calculate the filter response at each $v \in \V$, they are $K$-localized. In fact, they implement the weighted MA filtering in the form of FIR \cite{tremblay2018design}.
	
	Chebyshev polynomial approximation \cite{mason2002chebyshev} is one of the preferred methods  in signal processing due to their fast computation since they are generated via a recursion and not a convolution \cite{smith1997scientist}.
	The Chebyshev polynomial of the first kind $T_k(x)$ can be computed recursively $T_k(x) = 2x T_{k-1}(x) - T_{k-2}(x)$ where $T_0(x)=1$ and $T_1(x)=x$ \cite{mason2002chebyshev}.
	Thus, a filter $h$ can be approximated by a truncated expansion of Chebyshev polynomials $T_k$, up to order $K-1$. So, $\bm{X}$ can be filtered:

	\begin{equation}
		\bm{Y} = h \ast \bm{X} = h(\bm{L}) \bm{X} = \sum_{k=0}^{K-1} a_k T_k(\bm{\tilde{L}}) \bm{X}
	\end{equation}
	where $T_k(\bm{\tilde{L}}) \in \R^{n \times n}$ is the Chebyshev polynomial of order $k$ evaluated at the scaled Laplacian $\bm{\tilde{L}} = 2 \bm{L} / \lambda_{max} - \bm{I_n}$ \cite{defferrard2016convolutional} where $\bm{a} \in \R^K$ is a vector of Chebyshev coefficients.
	Full EVD can be omitted since this operation only requires the largest eigenvalue $\lambda_{max}$ which can be efficiently approximated by the power method \cite{follmer1998richard}.
	Although the MA type Chebyshev (CHEB) GFs are fast and localized, they often require high-degree polynomials to capture the graph's global structure.
	In fact, it restricts their ability to adapt sharp transitions in the frequency response due to the poor interpolation and extrapolation capabilities of high degree polynomials \cite{loukas2015distributed}.
		
	\subsection{Rational Graph Filters}
\begin{figure}[h!]
		\centering
		\includegraphics[width=0.97\linewidth]{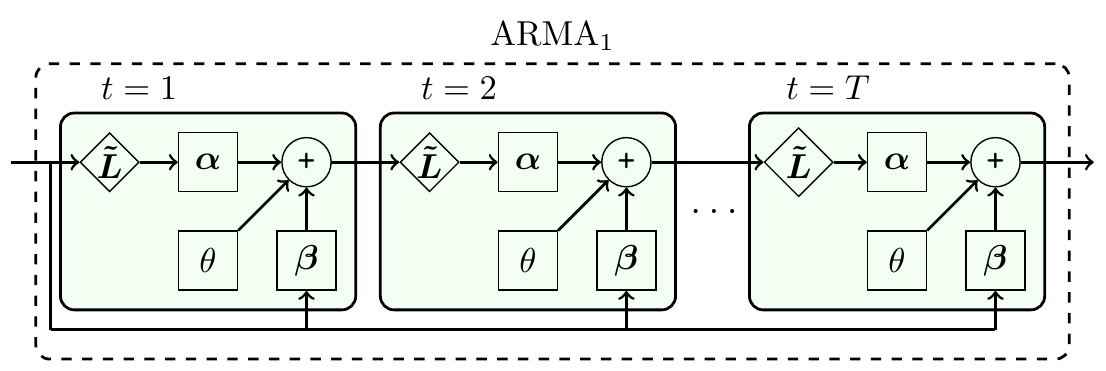}
		\caption{NN Implementation of ARMA$_1$ filter as a building block of ARMA$_K$ layer. In $T$ fixed iterations, an ARMA$_1$ block realizes eq.~(\ref{eq:y_arma_2})}
		\label{fig:arma1}
	\end{figure}

	To circumvent these problems, distributed IIR type ARMA GFs are proposed in \cite{loukas2015distributed, shi2015infinite}.
	They better approximate the sudden changes in the frequency response in comparison with the FIR type MA GFs due to their rational filter composition.
	A potential building block of $K$-order ARMA GFs may start with a first order recursive ARMA$_1$ filter:
	\begin{equation}\label{eq:y_arma_1}   
		\bm{Y}^{t+1} = a \bm{\tilde{L}} \bm{Y}^t + b \bm{X},
	\end{equation}
	where $\bm{Y}^t$ is the filter output at iteration $t$, $\bm{X}$ is the filter input, $a$ and $b$ are arbitrary coefficients, and modified Laplacian $\bm{\tilde{L}} = \frac{\lambda_{max} - \lambda_{min}}{2} \bm{I_n} - \bm{L}$ is a linear translation of $\bm{L}$ with same eigenvectors as $\bm{L}$ and shifted  eigenvalues $\tilde{\lambda}_n = \frac{\lambda_{max} - \lambda_{min}}{2} - \lambda_n$ relative to those of $\bm{L}$.
	According to the Theorem (1) in \cite{loukas2013think}, eq. (\ref{eq:y_arma_1}) converges regardless of $\bm{Y}^0$ and $\bm{L}$ values and its frequency response is given by $h_{ARMA_1}(\tilde{\lambda}_n) = \frac{b}{1 - a \tilde{\lambda}_n}$. 
	In fact, eq. (\ref{eq:y_arma_1}) provides a useful distributed filter realization \cite{shi2015infinite}.
	At each iteration $t$, each node $i$ revises its output $\bm{Y}_i^t \in \R^{n \times c_{out}}$ with a linear combination of its input $\bm{X_i} \in \R^{n \times c_{in}}$ and its adjacent nodes' outputs $\bm{Y}_j^{t-1}$, where $c_{in}$ and $c_{out}$ denote the number of channels in the input and the output tensors, respectively.
	It can be implemented as a NN layer if we unroll the recursion into $T$ fixed iterations:
	\begin{equation}\label{eq:y_arma_2}
	\bm{Y}^{t+1} = \bm{\tilde{L}} \bm{Y}^t \bm{\alpha} + \bm{X} \bm{\beta} + \bm{\theta},
	\end{equation}
	where $\bm{\alpha} \in \R^{c_{out} \times c_{out}}$, $\bm{\beta} \in \R^{c_{in} \times c_{out}}$, and $\bm{\theta} \in \R^{c_{out}}$ are trainable weights.
	Besides, since $0 \leq \lambda_{min} \leq \lambda_{max} \leq 2$, the  modified Laplacian can be simplified to $\bm{\tilde{L}} = \bm{I_n} - \bm{L}$ for $\lambda_{min}=0$, and $\lambda_{max}=2$ \cite{bianchi2021graph}. In Fig.~\ref{fig:arma1}, NN implementation of the ARMA$_1$ block which implements the eq.~(\ref{eq:y_arma_2}) in $T$ fixed iterations is depicted.
	ARMA$_1$'s $K$-order version ARMA$_K$ filter can be realized by averaging K parallel ARMA$_1$ filters with $\bm{Y} = \frac{1}{K}\sum_{k=1}^{K} \bm{Y}_K^T$ which leads to an ARMA$_K$ GF with a rational frequency response $h_{ARMA_{K}}(\tilde{\lambda}_n) = \sum_{k=1}^{K} \frac{b_k}{1 - a_k \tilde{\lambda}_n}$ with a $K-1$ and $K$ order polynomials in its numerator and denominator, respectively.
	For detailed analysis and justifications, please refer to \cite{loukas2015distributed, shi2015infinite, loukas2013think, isufi2016autoregressive, bianchi2021graph}.

	\subsection{Frequency Response of Polynomial and Rational GFs}
	To demonstrate the ARMA GFs better fit sharp changes in the frequency response compared to that of the CHEB GFs,
	we design two ideal GFs in equations~\eqref{eq:dagger} and~\eqref{eq:ddagger} for IEEE 118- and 300-bus test cases, respectively.
	\begin{equation} \label{eq:dagger}
		{h^\dagger}(\lambda) = 
		\begin{cases}
		1, & \frac{\lambda_{max}}{3} < \lambda  < \frac{2\lambda_{max}}{3} \\
		0, & \text{otherwise}
		\end{cases}
	\end{equation}
	\begin{equation} \label{eq:ddagger}
	{h^\ddagger}(\lambda) = 
	\begin{cases}
	1, & \lambda  < \frac{\lambda_{max}}{2} \\
	0, & \text{otherwise}
	\end{cases}
	\end{equation}
	Then, we investigate the approximating capability of the ARMA and the CHEB GFs by numerically analyzing their frequency responses.
	Note that similar results can be obtained by any other filters or test cases \cite{shi2015infinite}.	
	Let $\bm{x}, \bm{y} \in R^{n \times 1}$ denote the input and output of a GF $h(\lambda)$, respectively.
	Then, according to eq.~(\ref{eq:gsp}), empirical frequency response $\tilde{h}$ can be calculated by $\tilde{h}(\lambda_{i}) = \frac{\bm{u}_{i}^T \bm{y}}{ \bm{u}_{i}^T \bm{x}}$ \cite{loukas2015distributed}.
	Namely, each $\tilde{h}(\lambda_{i})$ represents how $\bm{u}_{i}$, corresponding to $\lambda_{i}$, ``scales'' $\bm{x}$ to obtain $\bm{y}$.

\begin{figure}[h!] 
	\centering
	\newcommand{\wdt}{0.6} 
	\subfloat[an ideal bandpass $h^\dagger$ for IEEE-118. \label{fig:118_filter}]{
		\centering
		\includegraphics[width=\wdt\linewidth]{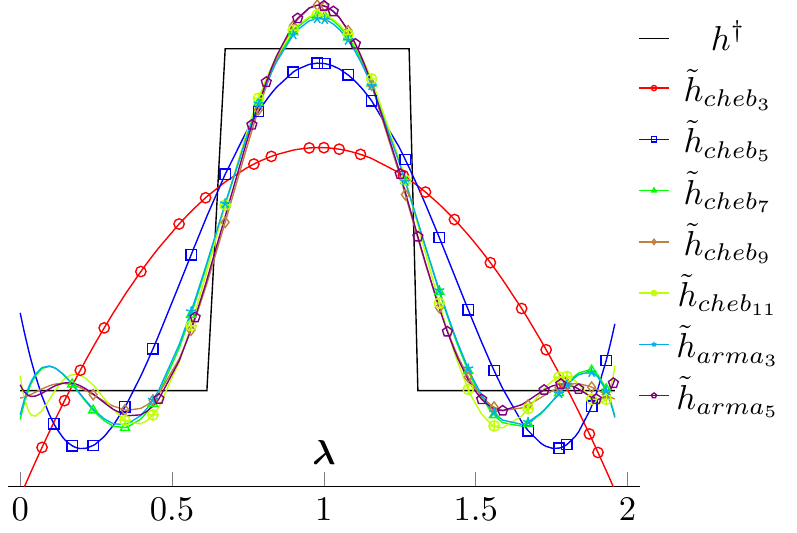}
	}
	\\
	\subfloat[an ideal lowpass $h^\ddagger$ for IEEE-300. \label{fig:300_filter}]{
		\centering
		\includegraphics[width=\wdt\linewidth]{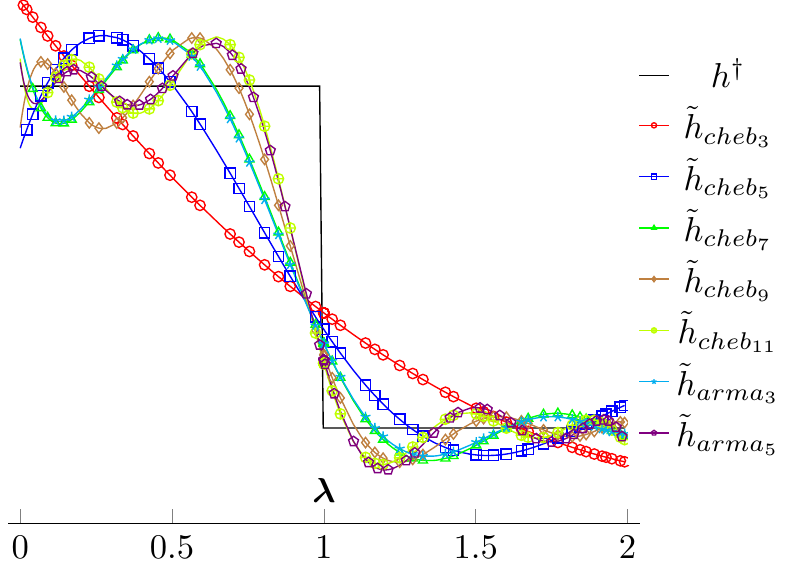}
	}
	\caption{Empirical frequency responses of CHEB and ARMA GFs when approximating ideal filters $h^\dagger$ and $h^\ddagger$ applied on IEEE 118- and 300-bus test systems, respectively. Compared to CHEB, ARMA better approximates the desired filter for the same $K$ (e.g. $\tilde{h}_{cheb_3}$ vs $\tilde{h}_{arma_3}$) and it requires a lower $K$ for the same level of approximation (e.g. $\tilde{h}_{cheb_{11}}$ vs $\tilde{h}_{arma_5}$).}
	\label{fig:filters}
	\end{figure}

	In order to obtain  $\tilde{h}(\lambda_{i})$ values, we first randomly generated $2^{16}$ $\bm{x}$s for the aforementioned systems from the normal distribution and filter them by $h^\dagger$ and $h^\ddagger$ using eq. (\ref{eq:gsp}) to obtain $\bm{y}$s.
	Then, a layer of CHEB and ARMA models with no activation function are trained in batches having $2^6$ samples of $\bm{x}$ and $\bm{y}$ values until there is no improvement.
	Next, $\tilde{h}(\lambda_{i})$ values are calculated for each $\bm{x}, \bm{y}$ tuple, averaged for smooth transitions, and plotted.
	As seen from Fig.~\ref{fig:filters}, due to their rational type frequency responses, ARMA GFs are more flexible to fit sudden changes for a fixed $K$ when compared to CHEB GFs having polynomial type frequency responses. This  constitutes the main motivation for selecting ARMA GFs for jointly detecting and localizing the FDIA in power grids.

	\subsection{Architecture of the Proposed Joint Detector \& Localizer}
\begin{figure}[h!]
	\centering
	\includegraphics[width=0.47\textwidth]{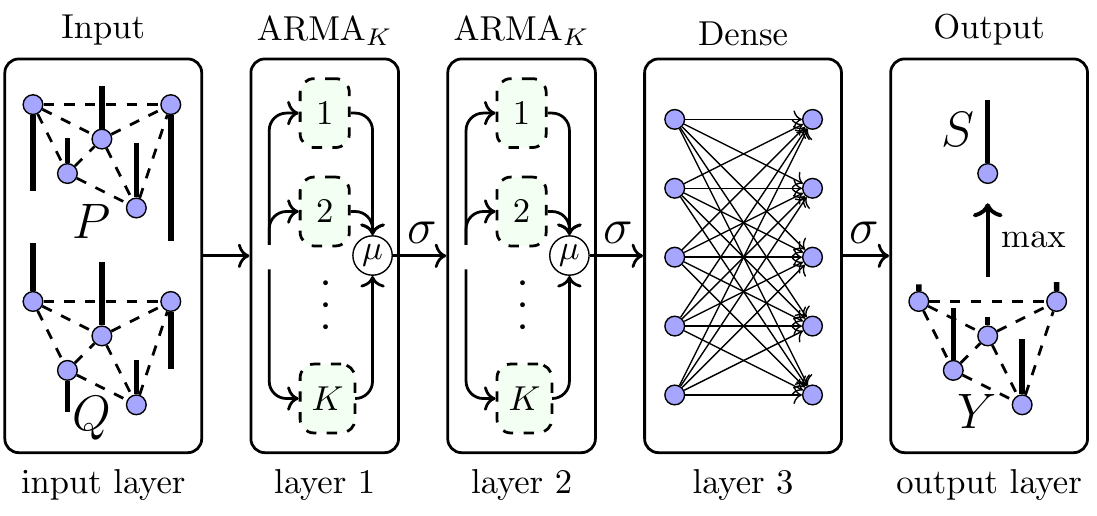}
	\caption{Architecture of the proposed ARMA GNN based detector and localizer with three hidden layers where each ARMA$_K$ layer consists of $K$ parallel ARMA$_1$. Each one of the $K$ dashed blocks in an ARMA$_K$ layer corresponds to an ARMA$_1$ block depicted with a dashed block in Fig.~\ref{fig:arma1}. While complex power injections $\bm{P},\bm{Q}$ and predicted attack probabilities $\bm{Y}, S$ at the node and graph level are visualized with thick bars at each node, activation and mean value functions are represented with $\sigma$ and $\mu$, respectively.}
	\label{fig:arhitecture}
	\end{figure}

	The proposed joint detector and localizer consists of one input layer to represent complex bus power injections, $L-1$ hidden ARMA$_K$ layers to extract spatial features, one dense layer to predict the probability of attack at each node, and one output layer to return an extra bit to indicate the probability of attack at the graph level. Its architecture is demonstrated in Fig.~\ref{fig:arhitecture} for $L=3$ with a small graph having $n=5$.

	In this multi-layer GNN model, the input tensor $[\bm{P_i}, \bm{Q_i}]$ is given by $\bm{X^0} \in \R^{n \times 2}$, the output tensor of hidden layer $l$ is denoted by $\bm{X^l} \in \R^{n \times c_l}$, and model outputs are denoted by $\bm{Y} \in \R^{n}$ and $S \in \R$ to indicate the location and the presence of the attack, respectively, where $c_l$ represents the number of channels in layer $l$ for $1 \leq l \leq L$.
	In particular, while an ARMA$_K$ layer takes $\bm{X}^{l-1} \in \R^{n \times c_{l-1}}$ as input and produces $\bm{X}^{l} \in \R^{n \times c_{l}}$ as output in layer $l$, the dense layer propagates the information to the whole graph and outputs the probability of the attack at the node level with $\bm{Y} \in \R^{n}$ for localization.
	Finally, the output layer  detects  the attack at the graph level by $S=\max(\bm{Y}) \in \R$ and outputs it with $\bm{Y}$. Note that the last ARMA$_K$ layer's output channel is selected as one in order to have one feature for each $v \in \V.$
	In addition, ReLU activation is used at the end of each ARMA$_K$ layer to increase the model's nonlinear modeling ability, whereas sigmoid is employed to transform the outputs to probabilities.

	\section{Experimental Results}\label{details} 
	\subsection{Data Generation}
	Due to the privacy concerns, there is no preexisting publicly available dataset to train and evaluate the proposed models against FDIA.
	Thus, researches use historical load profiles to mimic the timely deviations of the grids they simulate \cite{zhao2016detection, jevtic2018physics, hasnat2020detection, chaojun2015detecting, tan2016online, giraldo2016integrity}.
	We take the same approach based on the  historical load profile of NYISO~\cite{nyiso} to generate our dataset.
	As a first step, we download 5-minute intervals of the actual load profile of NYISO for July 2021 and interpolate them to increase the resolution to 1-minute.
	Next, we generate a realistic dataset following the Algorithm 1 in our previous work~\cite{boyaci2021graph} for the IEEE 57-, 118-, and 300-bus standard test cases using 1-minute interval load profile. Namely, for each timestamp, load values are distributed and scaled to buses proportional to their initial values, AC power flow algorithms are executed, and 1\% noisy power measurements are saved.
	
	To simulate the FDIA, we implement some of the frequently used FDIA generation algorithms in the literature such as data replay attacks ($A_r$) \cite{chaojun2015detecting, zhao2016detection}, data scale attacks ($A_s$) \cite{jevtic2018physics, hasnat2020detection}, and distribution-based ($A_d$) attacks \cite{ozay2015machine, yan2016detection} as well as our constrained optimization based FDIA ($A_o$) method explained thoroughly in~\cite{boyaci2021graph}.
	While $A_r$ simply changes a measurement $z_o^i$ with one of its previous values at $\tau$ back in time, $A_s$  multiply it with a number sampled from a uniform distribution ($\U$) between 0.9 and 1.1.
	In contrast, $A_d$ mimics the mean ($\mu$) and variance ($\sigma^2$) of the measurement by sampling from a normal distribution ($\N$) and $A_o$ solves a constrained optimization problem to maximize the changes in state variables while minimizing the changes in measurements.
	\begin{table}[h!]
		\centering
		\caption{Implemented FDIAs}
		\setlength{\tabcolsep}{3pt}
		\renewcommand{\arraystretch}{1.2}
		\newcolumntype{?}[1]{!{\vrule width #1}}
		\begin{tabular}{c ?{1pt} c ?{1pt} c }
			\textbf{FDIA type}     & \textbf{formulation} & \textbf{used in} \\ \specialrule{1pt}{1pt}{1pt}
			\textbf{optimization-based ($A_o$)} 	& Eq. (5) in \cite{boyaci2021graph} 		& \cite{boyaci2021graph} \\ \hline
			\textbf{data replay ($A_r$)	}		& $z_a^i(t) = z_o^i(t- \tau)$ 					& \cite{chaojun2015detecting, zhao2016detection} \\ \hline
			\textbf{distribution-based ($A_d$)}	& $z_a^i(t) = \N(\mu(z_o^i), \sigma^2(z_o^i))$ 	& \cite{ozay2015machine, yan2016detection} \\ \hline
			\textbf{data scale ($A_s$)}			& $z_a^i(t) = \U(0.9, 1.1) \times z_o^i(t)$ & \cite{jevtic2018physics, hasnat2020detection}  \\ 
		\end{tabular}
		\label{tab:fdia}
	\end{table}
	Implemented attacks types, their formulations and some works that have utilized them are given in Table~\ref{tab:fdia}.

	We shuffled the whole data to eliminate the seasonality, standardized it with a zero mean and a standard deviation of one to have a faster and smoother learning process, and split it into three sections: 2/3 for training, 1/6 for validating and hyper-parameter tuning, and 1/6 for testing the proposed models.
	In order to evaluate the performance of our method under unseen attack types, we arbitrarily selected $A_o$ and $A_d$ and included them in the training and validation splits. Test split, in contrast, includes all of the four FDIA methods given in Table~\ref{tab:fdia}.
	The number of honest samples are equalized with the number of malicious samples in each split to have a balanced classification problem as can be seen from Table \ref{tab:sample}.
	The final dataset assumes 60 samples/hour $\times$ 24 hour/day $\times$ 24 day = 34560 samples which consist of complex power measurements, complex bus voltages, and $n+1$ binary labels to indicate the true labels for each bus and the whole grid at each timestamp. 
	\begin{table}[h!]
		\centering
		\caption{Number of samples in each split}
		\setlength{\tabcolsep}{3pt}
		\renewcommand{\arraystretch}{1.2}
		\newcolumntype{?}[1]{!{\vrule width #1}}
		\begin{tabular}{c ?{1pt} c ?{1pt} c c c c ?{1pt} c}
			\textbf{split} & \textbf{non-attacked} & \textbf{$A_o$} & \textbf{$A_r$} & \textbf{$A_d$} & \textbf{$A_s$} & \textbf{total}  \\ \specialrule{1pt}{1pt}{1pt}
			\textbf{train} & 11520  & 5760 & 0   & 5760 &   0 & 23040 \\ \hline
			\textbf{validation} &  2880  & 1440 & 0   & 1440 &   0 &  5760 \\ \hline
			\textbf{test}  &  2880  &  720 & 720 & 720  & 720 &  5760 
		\end{tabular}
		\label{tab:sample}
	\end{table}
	
	\subsection{Feature Selection, Performance Metrics, and Training }
	To be able to rapidly detect and localize the attacks instead of waiting for $\bm{V_i}$ and $\bm{\theta_i}$ values at the output of the PSSE process, we employ power measurements as input features in our detectors. 
	From the power measurements, only $\bm{P_i}$ and $\bm{Q_i}$ values are fed to the models as seen from the input layer of Fig.~\ref{fig:arhitecture} since $\bm{P_i}+j\bm{Q_i} = \sum_{k \in \Omega_i} \bm{P_{ik}} +j\bm{Q_{ik}}$, node features can represent branch features as summation in their corresponding set of buses $\Omega_i$ connected to bus $i$.
	Besides, it is experimentally verified that utilizing $\bm{V_i}$ and $\bm{\theta_i}$ values along with $\bm{P_i}$ and $\bm{Q_i}$ does not increase the model performance due to tuples' high correlation. 
	PSSE and BDD modules, on the contrary, continue to receive every available measurement to operate.
	As for the weighted adjacency matrix we select $\bm{W} = |\bm{Y_{bus}}|$ to calculate $\bm{\tilde{L}}$ and feed the ARMA$_K$ layers where $\bm{Y_{bus}} \in \R^{n \times n}$ denotes nodal admittance matrix.
	
	For performance evaluation we use detection rate $DR = \frac{TP}{TP+FN}$, false alarm rate $FA = \frac{FP}{FP + TN}$, and $F1$ score $F1 = \frac{2*TP}{2TP + FP + FN}$, where $TP$, $FP$, $TN$, and $FN$ represent true positives, false positives, true negatives, and false negatives, respectively \cite{machine_learning}.
	In addition, to overcome the division by zero problem when there is no attack at all, we assumed $DR=1$, $FA=0$, and $F1=1$ if all the labels are correctly predicted as not attacked.
	Otherwise, even if there is one mismatch, we assign $DR=0$, $FA=1$, and $F1=0$. 

	All free unknown parameters defined in the model are computed by a multi-label supervised training using the binary cross-entropy loss.
	Training samples are fed into the model as mini batches of 256 samples with 256 maximum number of epochs.
	In addition, we employ early stopping criteria where 16 epochs are tolerated without any improvement less than $e^{-4}$ in the validation set's cross entropy loss.
	All the implementations were carried out in Python 3.8 using the Pandapower~\cite{pandapower}, Sklearn~\cite{sklearn}, t-SNE~\cite{tsne}, and Tensorflow~\cite{tensorflow} libraries on Intel i9-8950 HK CPU \@ 2.90GHz with NVIDIA GeForce RTX 2070 GPU.
	
	\subsection{Joint Detection and Localization Results} \label{sec:result}
	Since we take a data-driven approach in this work, we implement other existing data-driven approaches from the literature to compare with our method. To the best of our knowledge \cite{jevtic2018physics} is the only data-driven approach in the literature in which authors employ LSTM architecture to localize the FDIA. Thus, we trained an LSTM model with our dataset to compare the performances. In addition, although they are proposed for detection, we implement other available methods in the literature suitable for the multi-label classification task such as Decision Tree (DT) \cite{s98}, K-Nearest Neighbor (KNN) \cite{s99}, Multi Layer Perceptron  (MLP) \cite{s96}, Convolutional Neural Network (CNN) \cite{s99}, and Chebyshev GNN (CHEB) \cite{boyaci2021graph}. We train, validate and test these models similarly to the proposed model using our dataset as we do not have access to the data set of corresponding works.

	\begin{table}[h!] 
		\centering
		\setlength{\tabcolsep}{2pt}
		\newcolumntype{?}[1]{!{\vrule width #1}}
		\caption{Optimized model hyper-parameters.}
		\begin{tabular}{c?{1pt}c|c?{1pt}c?{1pt}c?{1pt}c}
			\textbf{model}  & \textbf{param} & \textbf{options} & \textbf{IEEE-57} & \textbf{IEEE-118} & \textbf{IEEE-300} \\
			\specialrule{1pt}{1pt}{1pt}
			
			\multirow{4}{*}{\textbf{DT}}  
			& criterion     & \{gini, entropy\}   	& gini    & entropy & entropy \\ \cline{2-6}
			& min. split    & \{2, 3, \dots, 8\} 	& 2       & 3       & 2       \\ \cline{2-6} 
			& max. depth    & \{1, 2, \dots, 64\}   & 60      & 64      & 64      \\ \cline{2-6}
			& features      & \{sqrt, log2\}      	& log2    & log2    & sqrt    \\ 
			\specialrule{1pt}{1pt}{1pt}
			
			\multirow{4}{*}{\textbf{KNN}}
			& algorithm     & \{ball, kd\}          & kd      & ball    & kd      \\ \cline{2-6} 
			& neighbors     & \{3, 5, 7, 9\}    & 3       & 3       & 3       \\ \cline{2-6} 
			& leaf size     & \{4, 5, \dots, 64\}   & 29      & 62      & 58      \\ \cline{2-6} 
			& p             & \{1, 2, 3, 4\}        & 2       & 1       & 1       \\
			\specialrule{1pt}{1pt}{1pt}
			
			\multirow{2}{*}{\textbf{MLP}}  
			& layers        & \{1, 2, 3\}           & 2       & 3       & 3       \\ \cline{2-6} 
			& units         & \{16, 32, 64, 128\}   & 32      & 32      & 128     \\
			\specialrule{1pt}{1pt}{1pt}
			
			\multirow{2}{*}{\textbf{LSTM}}  
			& layers        & \{1, 2, 3\}           & 3       & 3       & 2       \\ \cline{2-6} 
			& units         & \{16, 32, 64, 128\}   & 16      & 32      & 64      \\
			\specialrule{1pt}{1pt}{1pt}
			
			\multirow{3}{*}{\textbf{CNN}}  
			& layers        & \{1, 2, 3\}           & 3       & 2       & 3       \\ \cline{2-6} 
			& units         & \{16, 32, 64, 128\}   & 32      & 128     & 64      \\ \cline{2-6}
			& K             & \{2, 3, 4\}           & 3       & 4       & 3       \\
			\specialrule{1pt}{1pt}{1pt}  
			
			\multirow{3}{*}{\textbf{CHEB}}  
			& layers        & \{1, 2, 3\}           & 3       & 4       & 3       \\ \cline{2-6} 
			& units         & \{16, 32, 64, 128\}   & 64      & 32      & 64      \\ \cline{2-6}
			& K             & \{2, 3, 4\}           & 3       & 4       & 4       \\
			\specialrule{1pt}{1pt}{1pt} 
			
			\multirow{4}{*}{\textbf{ARMA}}  
			& layers        & \{1, 2, 3\}           & 3       & 2       & 3       \\ \cline{2-6} 
			& units         & \{16, 32, 64, 128\}   & 16      & 16      & 32      \\ \cline{2-6}
			& K             & \{2, 3, 4\}           & 2       & 3       & 3       \\ \cline{2-6}
			& iteration     & \{2, 3, 4, 5\}        & 4       & 5       & 5       \\ 
		\end{tabular}
		\label{tab:hpo}
	\end{table}

	Model hyper-parameters are tuned with Sklearn \cite{sklearn} and Keras-tuner \cite{kerastuner} Python libraries by using Bayesian optimization techniques.
	Models are trained on the training set and their hyper-parameters are optimized on the validation set for each IEEE test system for 250 trials.
	Finally, models with optimal parameters in terms of the validation set performance are saved and their results are presented for detection and localization. 
	Table~\ref{tab:hpo} shows the hyper-parameter set and the optimal hyper-parameters for each model and test system.
	
	\begin{table}[h]	
		\centering
		\caption{Detection results in $DR$, $FA$, and $F1$ percentages.}
		\setlength{\tabcolsep}{3pt}
		\renewcommand{\arraystretch}{1.2}
		\newcolumntype{?}[1]{!{\vrule width #1}}
		\begin{tabular}{c?{1pt}ccc?{1pt}ccc?{1pt}ccc}
			\textbf{System}
			& \multicolumn{3}{c?{1pt}}{\textbf{IEEE-57}} 
			& \multicolumn{3}{c?{1pt}}{\textbf{IEEE-118}} 
			& \multicolumn{3}{c}{\textbf{IEEE-300}}
			\\ \specialrule{1pt}{1pt}{1pt}
			\textbf{Metric}
			& \multicolumn{1}{c}{\bm{$DR$}} & \multicolumn{1}{c}{\bm{$FA$}} & \bm{$F1$}
			& \multicolumn{1}{c}{\bm{$DR$}} & \multicolumn{1}{c}{\bm{$FA$}} & \bm{$F1$}
			& \multicolumn{1}{c}{\bm{$DR$}} & \multicolumn{1}{c}{\bm{$FA$}} & \bm{$F1$}
			\\ \specialrule{1pt}{1pt}{1pt}
			\textbf{DT}
			& 89.55 &  5.45 & 91.84
			& 87.40 &  8.72 & 89.13
			& 89.69 &  9.38 & 90.11
			\\ \hline
			\textbf{KNN}
			& 19.41 & \textbf{0.07} & 32.49
			& 30.69 & \textbf{0.00} & 46.97
			& 16.67 & \textbf{0.00} & 28.57
			\\ \hline
			\textbf{MLP}
			& 95.07 & 0.31 & 97.32
			& 89.20 & 1.01 & 93.79
			& 82.74 & 1.63 & 89.76
			\\ \hline
			\textbf{LSTM}
			& 98.40 & 0.24 & 99.07 
			& 96.74 & 0.10 & 98.29 
			& 94.38 & 0.03 & 97.09 
			\\ \hline
			\textbf{CNN}
			& 99.79 & 0.28 & 99.76 
			& 98.47 & 0.45 & 99.01
			& 95.28 & \textbf{0.00} & 97.58
			\\ \hline
			\textbf{CHEB}
			& 99.65 & 0.28 & 99.69 
			& 97.99 & 0.45 & 98.76
			& 99.79 & 0.73 & 99.53
			\\ \hline
			\textbf{ARMA}
			& \textbf{99.90} & 0.28 & \textbf{99.81} 
			& \textbf{99.13} & 0.24 & \textbf{99.44}
			& \textbf{99.97} & 0.14 & \textbf{99.91} 
			\\ 
		\end{tabular}
		\label{tab:detection}
	\end{table}
	
	In Table~\ref{tab:detection}, detection performance of the optimized models for each test system is tabulated as percentages.
	For all test systems, although KNN yields the best $FA$ rate, its $F1$ scores are not satisfactory since it gives the lowest $DR$. 
	ARMA, in contrast, reaches the best $F1$ scores with $99.81\%$, $99.44\%$, and $99.91\%$, due to its high $DR$ with $99.90\%$, $99.13\%$, and $99.97\%$ and low $FA$ rate with $0.28\%$, $0.24\%$, and $0.14\%$ for 57-, 118-, and 300-bus systems, respectively.
	Although detection results are close to each other in terms of $F1$ scores for some models such as CHEB and ARMA, CHEB yields almost two and five times $FA$ rate for IEEE 118- and 300- bus systems, respectively. 
	Nevertheless, detection considers the attacks at the grid level and any intrusion to a bus in the grid is regarded as an attack.
	Thus, bus level localization is required to determine the exact place of the attack.
	
	Since localization is a multi-label classification problem, we evaluate it in both possible ways: (i) sample-wise (SW) evaluation yields $b$ metrics where each one of $b$ samples at a fixed time-step is treated individually along the buses, and (ii) node-wise (NW) evaluation yields $n$ metrics where each one of $n$ buses is evaluated separately along the samples.
	Therefore, in order to better assess the models, we visualize and analyze the distributions of SW and NW localization results in $F1$ percentages by box plots and ratio of items satisfying some specified thresholds.
	Box plots helps us to visualize the distribution by drawing first ($Q_1$, $25\textsuperscript{th}$ percentile), second ($Q_2$, $50\textsuperscript{th}$ percentile or median), and third ($Q_3$, $75\textsuperscript{th}$ percentile) quartiles, lower ($LW=Q_1-1.5 \times (Q_3-Q_1)$) and upper ($UW=Q_3+1.5 \times (Q_3-Q_1) $) whiskers and outliers \cite{tukey1977exploratory}.
	In addition, the ratio of the samples or buses satisfying some thresholds provides quantifiable metrics to assess model performances.
	For instance, the percentage of samples (buses) having $F1 \le 5\%$ or $F1 \ge 95\%$ in SW (NW) evaluation can be used to measure the ratio of ``unacceptable'' and ``acceptable'' samples in the distributions, respectively.

\begin{figure}[h!] 
		\centering
		\newcommand{\wdt}{0.97} 
		\subfloat[IEEE-57. $Q_2=100\%$ except DT, $Q_3=100\%$ for all. \label{fig:swe_057}]{
			\centering
			\includegraphics[width=\wdt\linewidth]{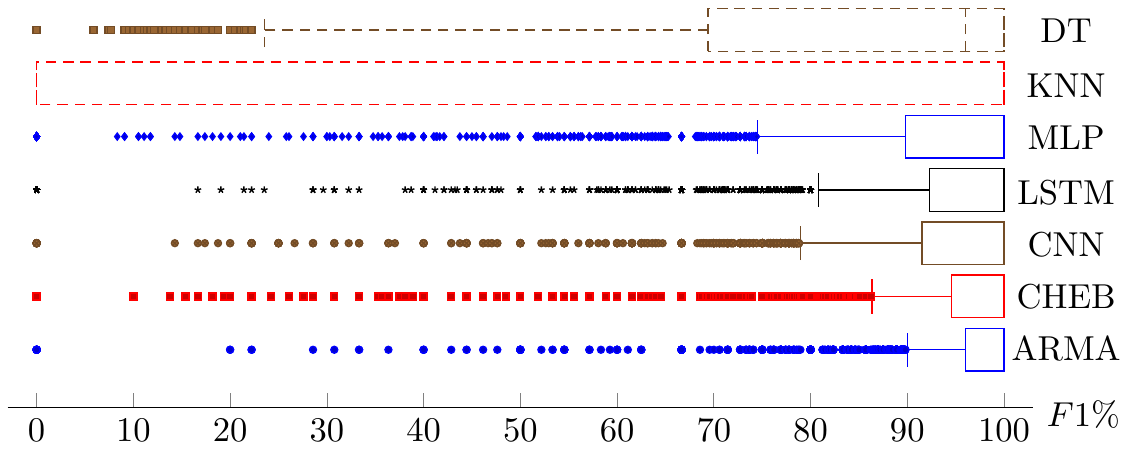}
		}
		\\
		\subfloat[IEEE-118. $Q_2=100\%$ except DT, $Q_3=100\%$ for all. \label{fig:swe_118}]{
			\centering
			\includegraphics[width=\wdt\linewidth]{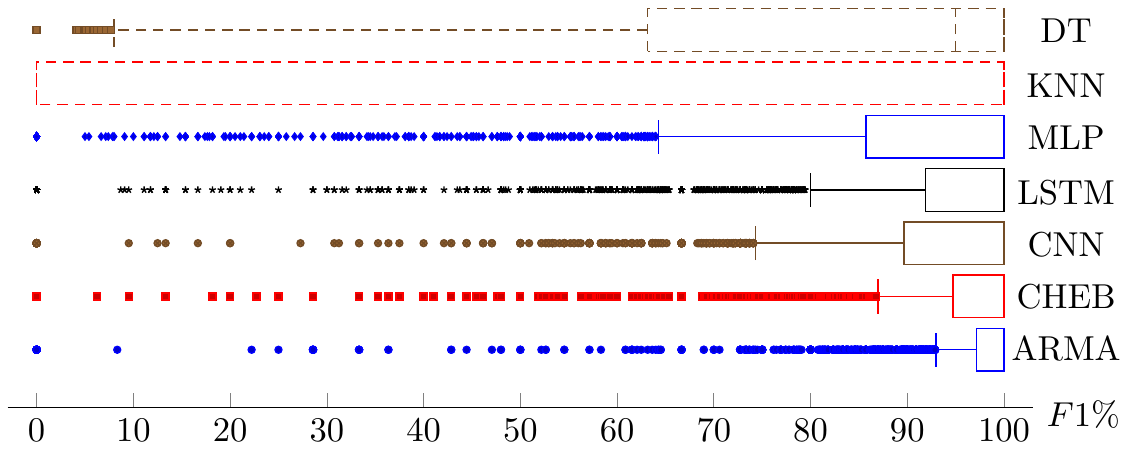}
		}
		\\
		\subfloat[IEEE-300. $Q_2=100\%$ except DT and MLP, $Q_3=100\%$ for all. \label{fig:swe_300}]{
			\centering
			\includegraphics[width=\wdt\linewidth]{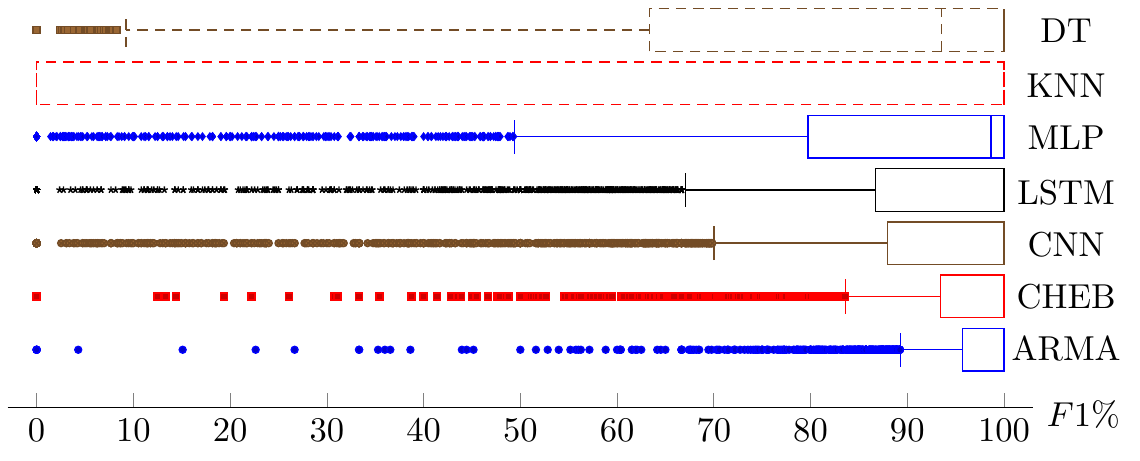}
		}\\
		\subfloat[Sample percentages having $F1 \le 5\%$ and $F1 \ge 95\%$. \label{fig:swe_tab}]{
			\centering
			\includegraphics[width=0.80\linewidth]{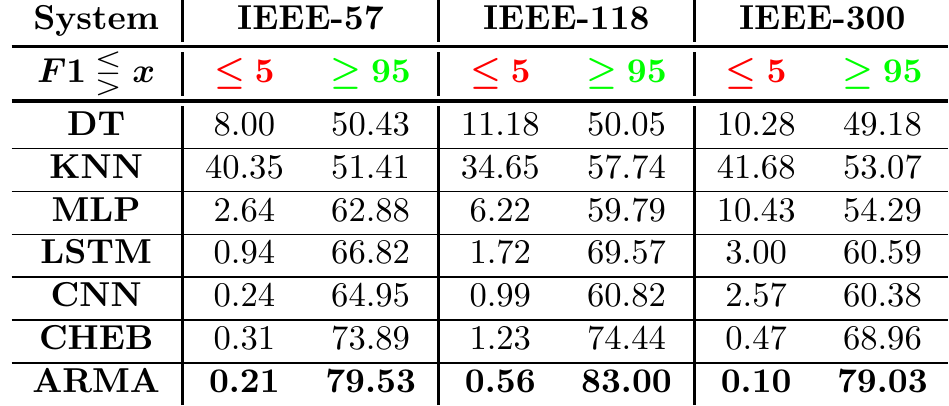}
		}\\
		\caption{Distribution of $F1$ scores for sample wise evaluation of localization.}
		\label{fig:localization_swe}
	\end{figure}

	SW localization results are given in Fig.~\ref{fig:localization_swe}.
	Since the distributions are highly left skewed, the median ($Q_2$) values can overlap with $Q_3$.
	In specific, $Q_2 = Q_3 = UW = 100\%$ except the MLP for IEEE-300 and DT for all systems.
	This is not surprising because 50\% of the samples are not attacked and it is relatively easy to predict them as not attacked for each bus.
	$Q1$ and $LW$, in contrast, vary for each model and test system.
	For example, $LW = Q_1 = 0$ in all test systems for KNN model which shows that $F1 = 0\%$ for more than 1/4 of the samples for that model.
	Although DT yields better results compared to KNN, its results are unsatisfactory:  its $Q_1$ values are $69.39\%$, $63.16\%$, and $63.33\%$, for 57-, 118-and 300-bus test systems, respectively. 
	Models from the NN family better localize the attacks compared to the classical ML approaches.	
	For example, their $Q1$ values are greater than $79.74\%$ for all test systems.
	Namely, in at least 3/4 of the samples, attacked buses are correctly labeled with $F1$ score deviating between $79.74\%$ and $100\%$.
	To better compare the model performances, the percentages of the samples having $F1 \le 5\%$ and $F1 \ge 95\%$ are given in Fig. \ref{fig:swe_tab} for each model and test system.
	ARMA gives the best results in all test cases: while the sample percentages for $F1 \le 5\%$ are calculated as $0.21\%$, $0.56\%$, and $0.10\%$, sample percentages with $F1 \ge 95\%$ are measured as $79.53\%$, $83.00\%$, and $79.03\%$ for IEEE 57-, 118-, and 300-bus test systems, respectively. 
	Its ``acceptable'' ($F1 \ge 95\%$) percentages are $5.64\%$, $8.56\%$, and $10.07\%$ greater than the second best model CHEB in SW localization, for IEEE 57-, 118-, and 300-bus test cases, respectively. 
	
	\begin{figure}[h!] 
		\centering
		\newcommand{\wdt}{0.97} 
		\subfloat[IEEE-57. $Q_3 = 100\%$ for all. \label{fig:nwe_057}]{
			\centering
			\includegraphics[width=\wdt\linewidth]{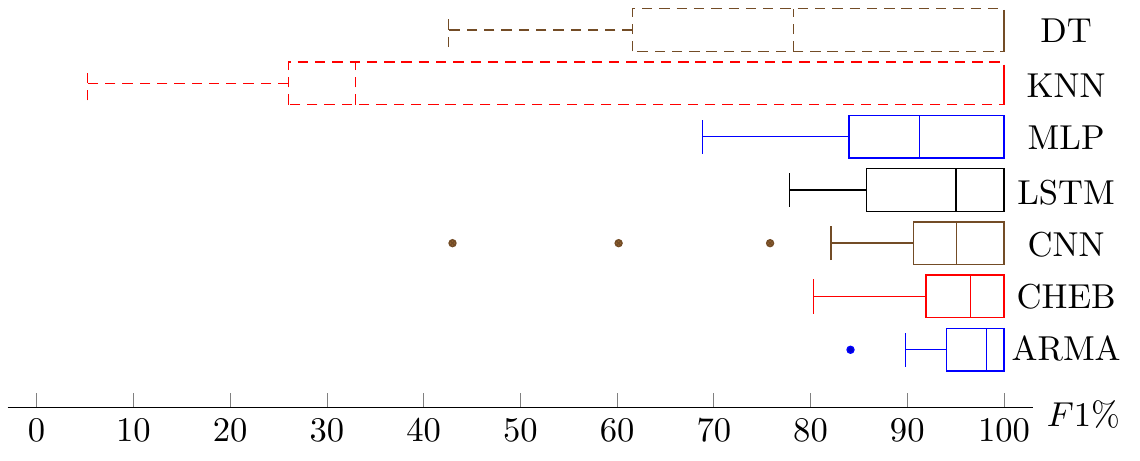}
		}\\
		\subfloat[IEEE-118. $Q_2 = Q_3 = 100\%$ for all. \label{fig:nwe_118}]{
			\centering
			\includegraphics[width=\wdt\linewidth]{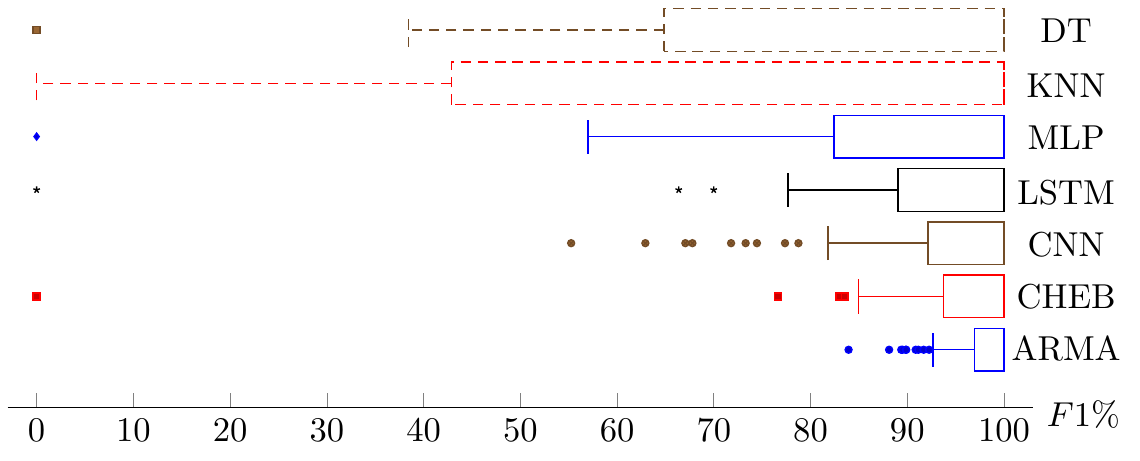}
		}\\
		\subfloat[IEEE-300. $Q_3 = 100\%$ for all. \label{fig:nwe_300}]{
			\centering
			\includegraphics[width=\wdt\linewidth]{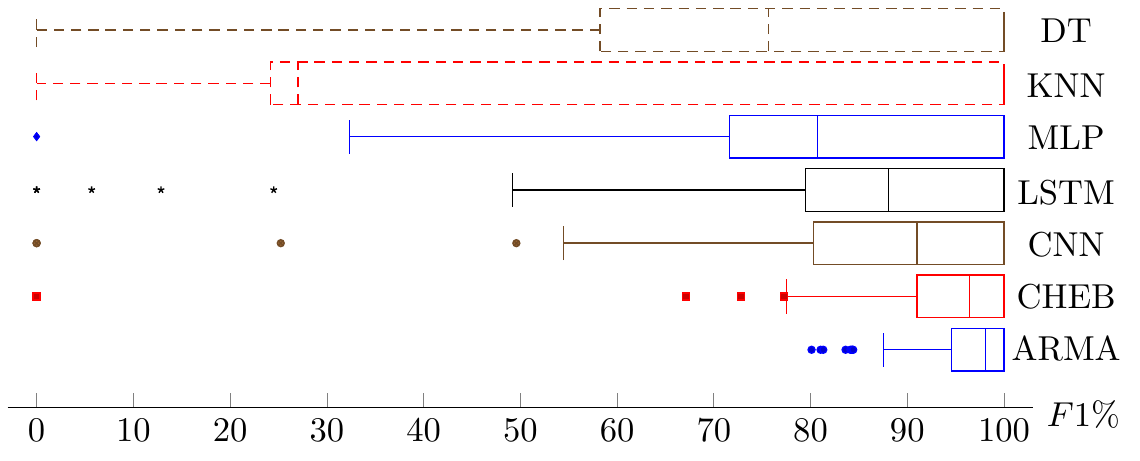}
		}\\
		\subfloat[Bus percentages having $F1 \le 5\%$ and $F1 \ge 95\%$. \label{fig:nwe_tab}]{
			\centering
			\includegraphics[width=0.80\linewidth]{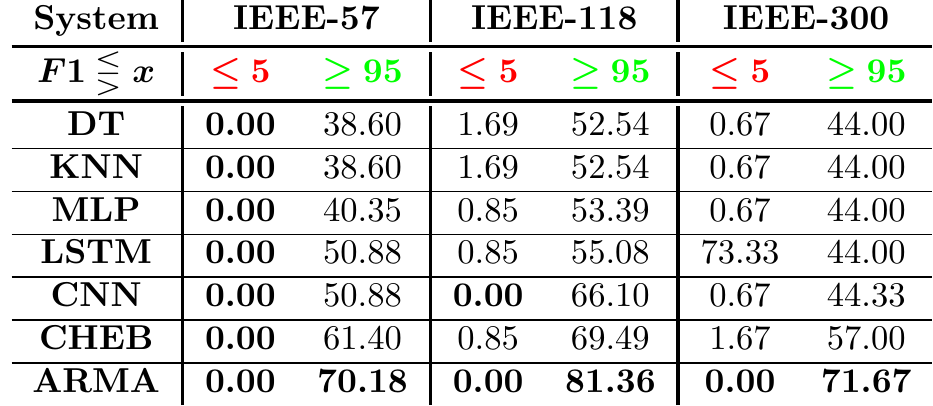}
		}\\
		\caption{Distribution of $F1$ scores for node wise evaluation of localization.}
		\label{fig:localization_nwe}
	\end{figure}
	
	In Fig.~\ref{fig:localization_nwe}, the distribution of $F1$ scores for NW evaluation is depicted.
	Due to the largely left skewed distributions, the median ($Q_2$) values may overlap with $Q_3$.
	Specifally, $Q_3 = 100\%$ for all the models and systems, and $Q2=100\%$ for all the models in IEEE-118.
	Similar to the SW evaluation, performance of DT and KNN is poor: their $Q_1$ values deviate between $24.19\%$ and $64.86\%$.
	MLP, LSTM, and CNN provide better results compared to DT and KNN. Nevertheless, they are subject to some outliers at $0\%$ which means there are some buses that are always mislabeled at each timestamp.
	The only model that can localize the FDIA for each bus with at least $80\%$ $F1$ score is ARMA.
	Namely, for all the test systems, the ARMA model can determine the location of an FDIA attack for all buses with $F1$ score greater than $80\%$. 
	Fig.~\ref{fig:nwe_tab} presents the percentages of buses satisfying $F1 \le 5\%$ and $F1 \ge 95\%$ levels.
	For all test systems, only ARMA model has $0\%$ with $F1 \le 5\%$ success level which means only ARMA model doesn't yield any ``unacceptable'' bus localization performance.
	In comparison, one bus in IEEE 118- and 5 buses in IEEE 300-bus systems always have $F1$ score less than $5\%$ in all timestamps for the second best CHEB model.
	For the $F1 \ge 95\%$ threshold, only ARMA model can surpass the $70\%$ level for each test systems and it outperforms the second best model CHEB by $8.78\%$, $11.87\%$, and $14.67\%$ for the $95\%$ $F1$ threshold level in NW localization for IEEE 57-, 118-, and 300-bus systems, respectively.

	\subsection{Joint Detection and Localization Times} \label{sec:time}
	We measure the elapsed time during model's joint detection and localization process for each sample in the test set, calculate the mean values, and tabulate them in Table~\ref{tab:time}.	
	\begin{table}[h!] 
		\centering
		\caption{Joint detection and localization times in milliseconds.}
		\setlength{\tabcolsep}{3pt}
		\renewcommand{\arraystretch}{1.2}
		\newcolumntype{?}[1]{!{\vrule width #1}}
		\begin{tabular}{c ?{1pt} c| c | c | c | c | c | c}
			\textbf{model} & \textbf{DT} & \textbf{KNN} & \textbf{MLP} & \textbf{LSTM} & \textbf{CNN} & \textbf{CHEB} & \textbf{ARMA} \\ \specialrule{1pt}{1pt}{1pt}
			\textbf{IEEE-57}  & 0.18 & 147.78 & 1.42 & 16.85 & 2.64 & 2.24 & 2.76 \\ \hline
			\textbf{IEEE-118} & 0.29 & 327.62 & 1.44 & 35.19 & 2.67 & 2.54 & 2.81 \\ \hline
			\textbf{IEEE-300} & 0.69 & 836.52 & 1.50 & 99.78 & 2.73 & 2.71 & 2.94
		\end{tabular}
		\label{tab:time}
	\end{table}
	Clearly, detection times of KNN are not satisfactory: it can take more than 0.8 second to respond.
	It is due to the fact that in KNN each new sample has to be compared with others for proximity calculation.
	LSTM, in contrast, provides better results compared to the KNN.
	Nevertheless, its highly complex recurrent architecture can delay its output almost 0.1 second for IEEE-300, which may limit its application in a real time scenario.
	All the other models including DT, MLP, CNN, CHEB, and ARMA provide reasonable detection times for a real time application: for all test system their response time is less than 3 milliseconds.
	Among them DT provides the best detection times with values under 0.7 milliseconds; yet, its poor detection and localization performance hinders its suitability as a reliable method.

	\subsection{Visualization of intermediate layers with t-SNE} \label{sec:tsne}
	
	To compare the proposed model with the existing approaches, we analyze and visualize the multidimensional data processed by the intermediate layers of the proposed models.
	Nevertheless, the high dimensionality of the data severely limits examining them.
	Besides, examining a specific feature of a bus does not provide enough information to fully comprehend how the model processes the grid.
	Thus, we transform the layer outputs by using the t-distributed stochastic neighbor embedding (t-SNE), which is a nonlinear dimensionality reduction technique to visualize the high dimensional data in two or three dimensional spaces \cite{tsne}.
	By iteratively minimizing the Kullback-Leibler divergence between the probability distributions representing the sample similarities in the original and mapped spaces, it projects samples into the low dimensional space.
	Thus, it preserves  the structure of the data and enables visualization of the data in a lower dimension \cite{tsne}.
	
\begin{figure}[h!]
		\centering
		\newcommand{\f}{tsne}
		\newcommand{\n}{300}
		\newcommand{\wdt}{0.47} 
		\subfloat[Embedded $\bm{X}$ in 2D\label{fig:tsne_in}]{
			\centering
			\includegraphics[width=\wdt\linewidth]{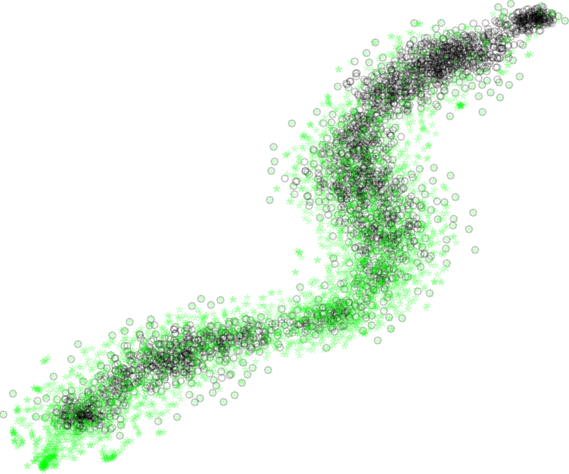}
		}
		\subfloat[Embedded $\bm{Y}$ in 2D\label{fig:tsne_out}]{
			\centering
			\includegraphics[width=\wdt\linewidth]{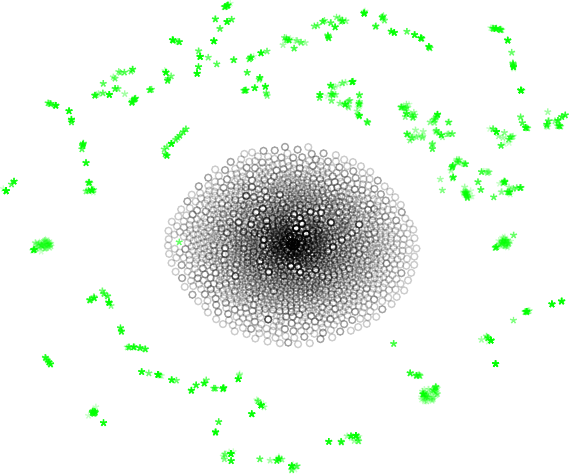}
		}
		\caption{Embedded input and output data for IEEE-300 bus system where attacked and non-attacked samples are depicted with green stars and black circles, respectively.}
		\label{fig:tsne_inout}
	\end{figure}

	Due to space limitations, only models trained for the IEEE-300 bus system are analyzed in two dimensions (2D) with test data having $5,760$ samples.
	Embedding of input data $\bm{[P,Q]=X} \in \R^{300 \times 2}$ is plotted in Fig.~\ref{fig:tsne_in}, where a dominating daily profile can be seen from the smooth transition from the lower left to the upper right samples depicted with green stars (attacked) and black circles (non-attacked).
	Moreover, the close proximity between attacked and non-attacked samples indicates that the attacked samples preserve similarity to their non-attacked samples. 
	Fig.~\ref{fig:tsne_out} shows the embedding of true output $\bm{Y} \in \R^{300}$ where non-attacked samples clustered in the middle and attacked samples are scattered around them.
	This is not surprising since non-attacked samples are all formed from 0s and attacked samples include 1s in their corresponding labels to indicate the attacked bus. 

\begin{figure*}[ht!]	
		\centering
		\newcommand{\f}{tsne}  
		\newcommand{\n}{300}  
		\newcommand{\wdt}{0.18} 
		\setlength{\tabcolsep}{2pt}
		\newcolumntype{?}[1]{!{\vrule width #1}}
		\begin{tabular}{c?{1pt}c?{1pt}c?{1pt}c?{1pt}c?{1pt}c}
			$\bm{l}$ &
			\begin{minipage}[c]{\wdt\textwidth}
				\centering \textbf{MLP} \\ \textcolor{green}{2383}, \textcolor{blue}{47}, \textcolor{red}{497}, \textcolor{black}{2833}
			\end{minipage} &
			\begin{minipage}[c]{\wdt\textwidth}
				\centering \textbf{LSTM} \\ \textcolor{green}{2718}, \textcolor{blue}{1}, \textcolor{red}{162}, \textcolor{black}{2879}
			\end{minipage} &
			\begin{minipage}[c]{\wdt\textwidth}
				\centering \textbf{CNN} \\ \textcolor{green}{2744}, \textcolor{blue}{0}, \textcolor{red}{136}, \textcolor{black}{2880}
			\end{minipage} &
			\begin{minipage}[c]{\wdt\textwidth}
				\centering \textbf{CHEB}\\ \textcolor{green}{2874}, \textcolor{blue}{21}, \textcolor{red}{6}, \textcolor{black}{2859}
			\end{minipage} &
			\begin{minipage}[c]{\wdt\textwidth}
				\centering \textbf{ARMA}\\ \textcolor{green}{2879}, \textcolor{blue}{4}, \textcolor{red}{1}, \textcolor{black}{2876}
			\end{minipage}
			\\ \specialrule{1pt}{1pt}{1pt} 
			$l_1$ &
			\begin{minipage}[c]{\wdt\textwidth}
				\includegraphics[width=\linewidth]{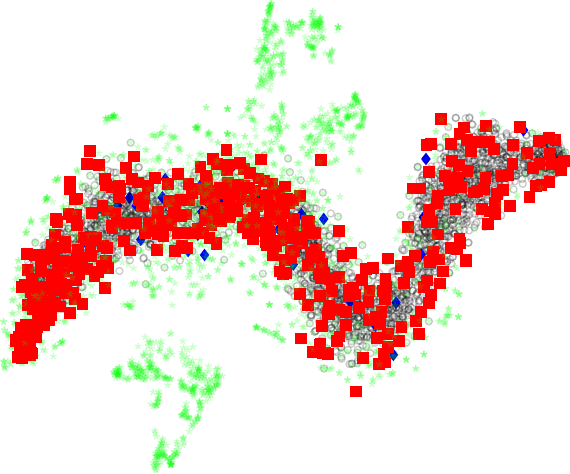}
			\end{minipage} &
			\begin{minipage}[c]{\wdt\textwidth}
				\includegraphics[width=\linewidth]{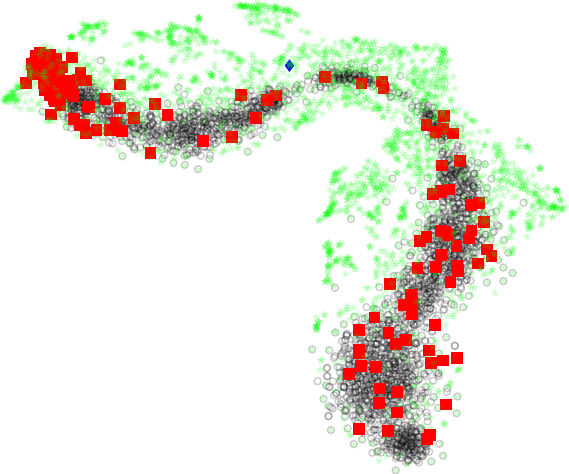}
			\end{minipage} &
			\begin{minipage}[c]{\wdt\textwidth}
				\includegraphics[width=\linewidth]{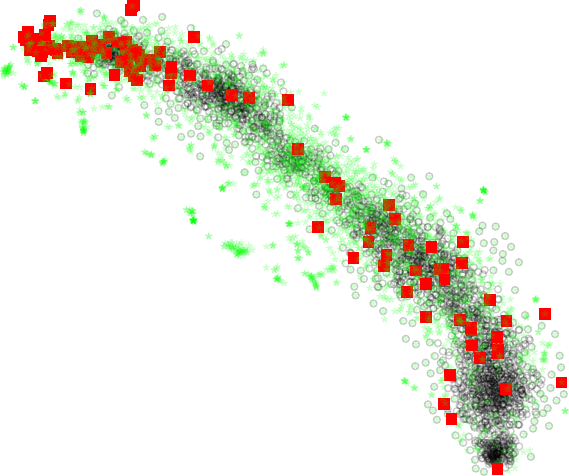}
			\end{minipage} &
			\begin{minipage}[c]{\wdt\textwidth}
				\includegraphics[width=\linewidth]{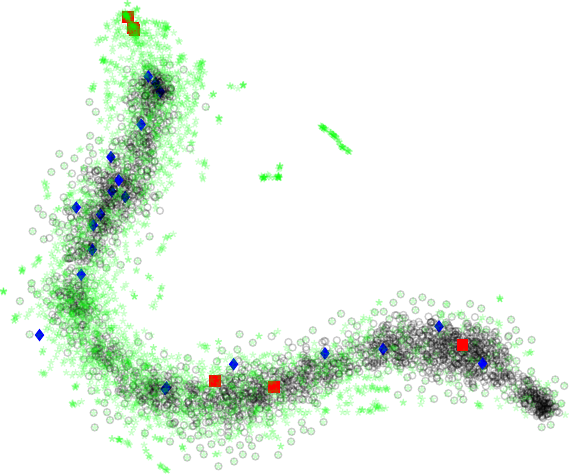}
			\end{minipage} &
			\begin{minipage}[c]{\wdt\textwidth}
				\includegraphics[width=\linewidth]{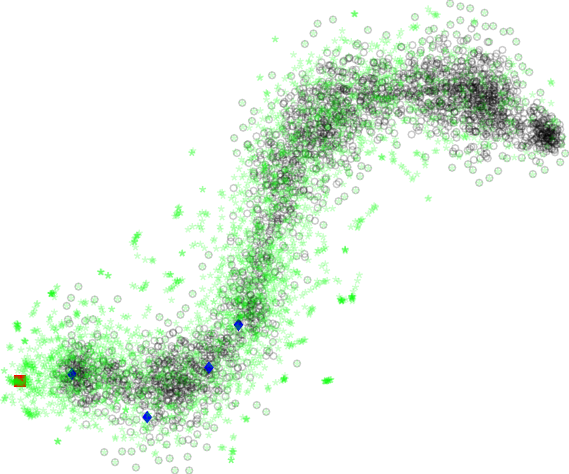}
			\end{minipage}
			\\ \hline  
			$l_2$ &
			\begin{minipage}[c]{\wdt\textwidth}
				\includegraphics[width=\linewidth]{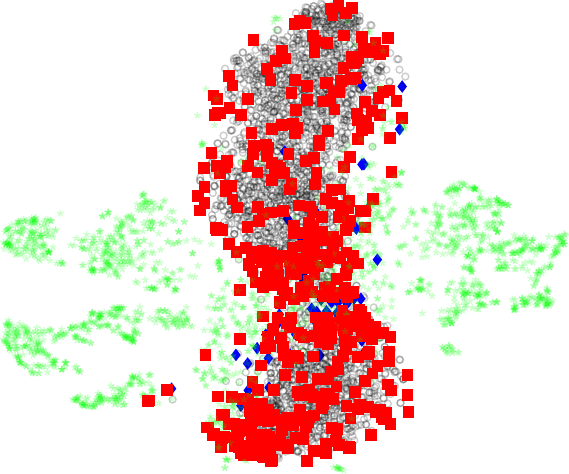}
			\end{minipage} &
			\begin{minipage}[c]{\wdt\textwidth}
				\includegraphics[width=\linewidth]{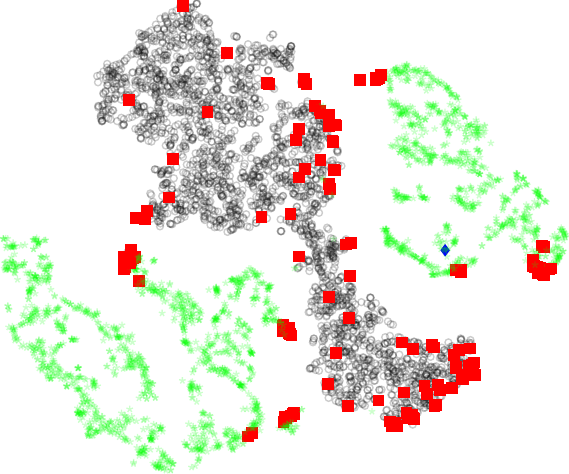}
			\end{minipage} &
			\begin{minipage}[c]{\wdt\textwidth}
				\includegraphics[width=\linewidth]{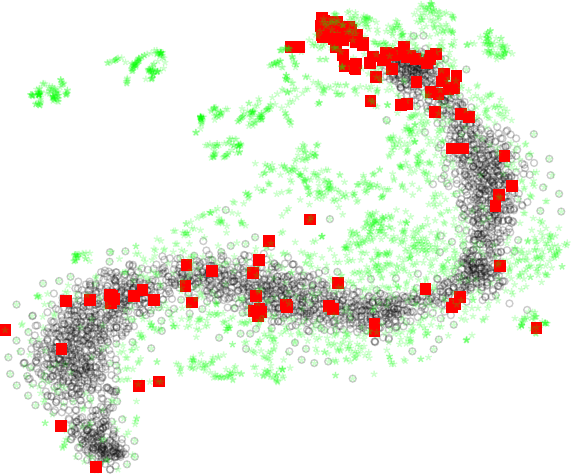}
			\end{minipage} &
			\begin{minipage}[c]{\wdt\textwidth}
				\includegraphics[width=\linewidth]{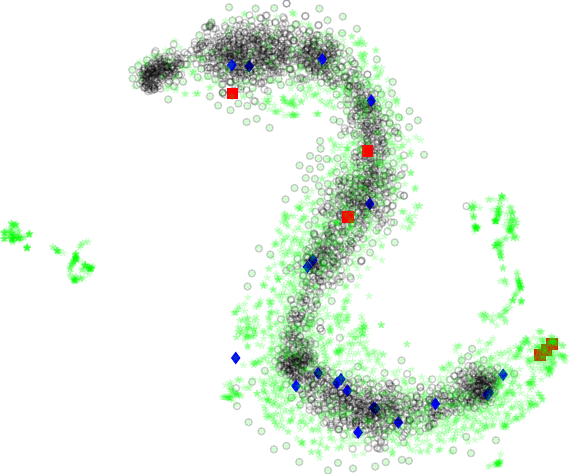}
			\end{minipage} &
			\begin{minipage}[c]{\wdt\textwidth}
				\includegraphics[width=\linewidth]{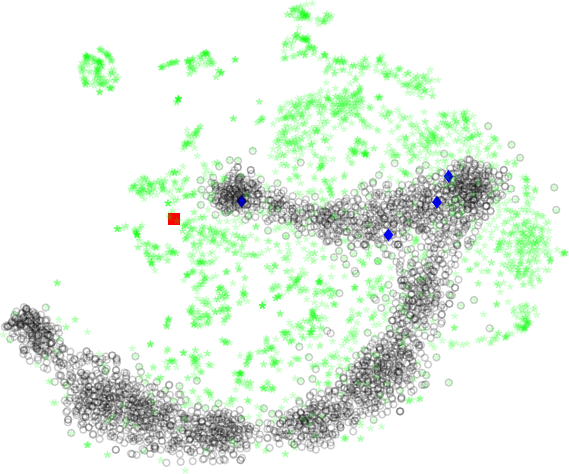}
			\end{minipage}
			\\ \hline  
			$l_3$ &
			\begin{minipage}[c]{\wdt\textwidth}
				\includegraphics[width=\linewidth]{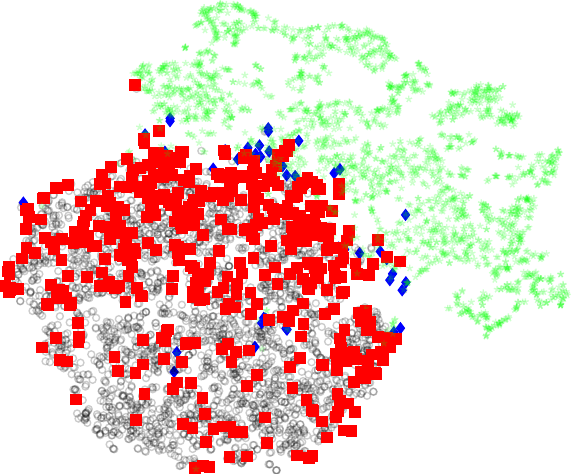}
			\end{minipage} &
			\begin{minipage}[c]{\wdt\textwidth}
				\includegraphics[width=\linewidth]{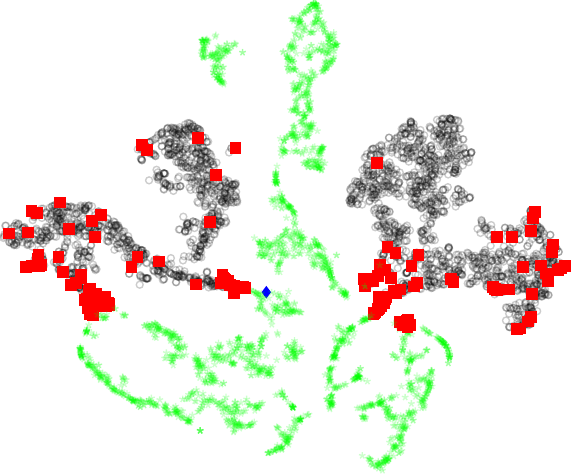}
			\end{minipage} &
			\begin{minipage}[c]{\wdt\textwidth}
				\includegraphics[width=\linewidth]{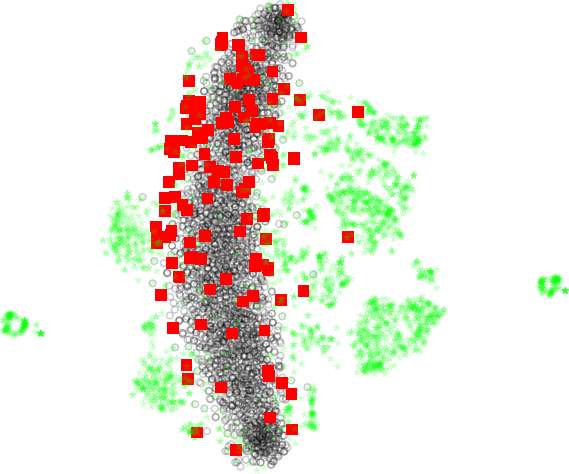}
			\end{minipage} &
			\begin{minipage}[c]{\wdt\textwidth}
				\includegraphics[width=\linewidth]{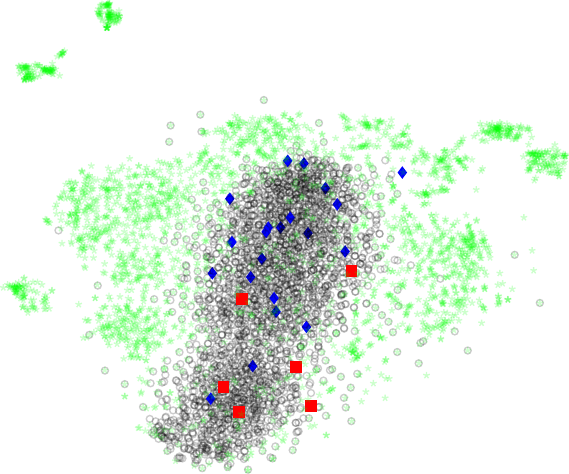}
			\end{minipage} &
			\begin{minipage}[c]{\wdt\textwidth}
				\includegraphics[width=\linewidth]{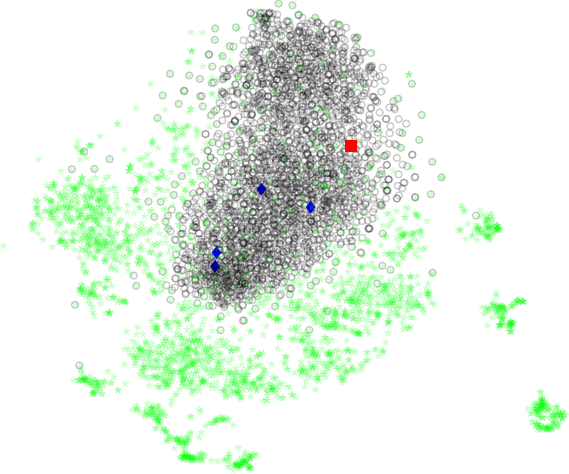}
			\end{minipage}
			\\ \hline  
			$l_4$ &
			\begin{minipage}[c]{\wdt\textwidth}
				\includegraphics[width=\linewidth]{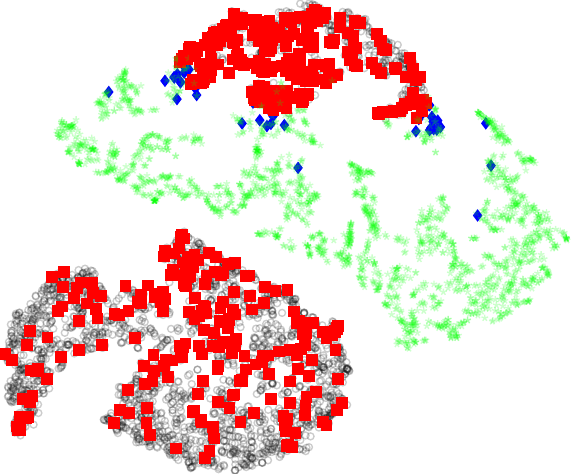}
			\end{minipage} &
			\begin{minipage}[c]{\wdt\textwidth}
				\includegraphics[width=\linewidth]{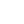}
			\end{minipage} &
			\begin{minipage}[c]{\wdt\textwidth}
				\includegraphics[width=\linewidth]{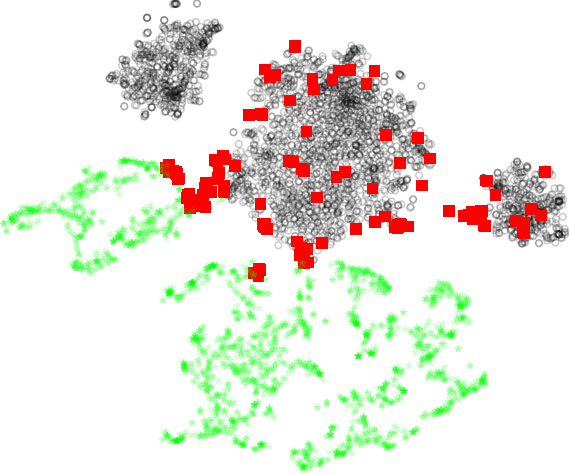}
			\end{minipage} &
			\begin{minipage}[c]{\wdt\textwidth}
				\includegraphics[width=\linewidth]{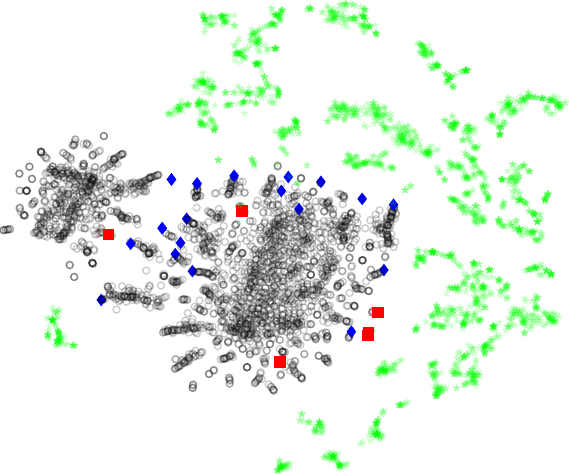}
			\end{minipage} &
			\begin{minipage}[c]{\wdt\textwidth}
				\includegraphics[width=\linewidth]{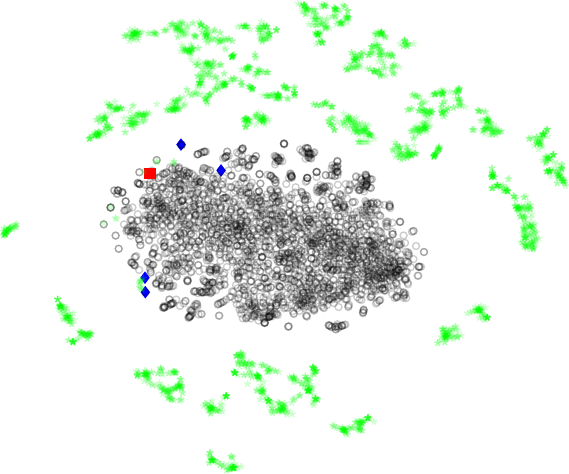}
			\end{minipage}
		\end{tabular}
		\caption{t-SNE embedding of model's layers to visualize the attack detection where true input and output data are given in Fig.~\ref{fig:tsne_inout}. For each model and each layer $l$, output of the model is embedded in 2D using t-SNE. Since t-SNE preserves the structure of the high dimensional data, models' transformation can be visualized and  compared in a lower dimension, such as 2D. Note that due to its topology aware ARMA graph filtering, the proposed model better classifies samples by converging to  the true output depicted in Fig.~\ref{fig:tsne_out}. As a consequence, it yields the minimum number of FP and FN compared to other models.}
		\label{fig:tsne_feature}
	\end{figure*}

	Fig.~\ref{fig:tsne_feature} demonstrates the embedding of layer $l$'s output where each model takes $\bm{X}$ (Fig.~\ref{fig:tsne_in}) as input, transforms it in the hidden layers, and tries to predict $\bm{Y}$ (Fig.~\ref{fig:tsne_out}) as output. The number of TP (green stars), FP (blue diamonds), FN (red squares), and TN (black circles) samples are given under the model name for easy reference.
	MLP clearly falls behind the other approaches due to the FNs scattered all around.
	For instance, unlike other approaches, MLP misses easily detectable attacked samples in $l_2$ very close to the TP cluster and it maps many FNs nearby to the TNs placed at the lower part of its last layer.
	LSTM, in contrast to the MLP, reduces FN and FP samples.
	However, in $l_2$, it falsely maps many attacked samples adjacent to the non-attacked samples which yields a high number of FNs.
	In addition, like the MLP, it falsely predicts multiple non-attacked clusters in its final layer.

	Contrarily, CNN is one of the best models in terms of FP number.
	Yet, it ``destroys'' the structure of data in $l_1$ which brings a significant number of FNs.
	We believe it is due to the fact that CNN tries to capture the correlations of non-Euclidean data in an Euclidean space and samples from different classes may look the same in that space.
	Due to their inherent graph architecture, CHEB and ARMA yield  better results since they both consider the ``structure'' of the data within $\bm{X}$ in their graph convolutional layers.
	However, CHEB misses 5 more attacks and yields more than 5 times FP samples compared to the ARMA.	
	For instance, many non-attacked samples in $l_4$ are falsely regarded as an attack due to close mapping to an attacked cluster.
	Conversely, our proposed model gives only 4 FP and 1 FN due to its rational graph convolutional filters that provide more flexible frequency responses.
	Note that no sample is mapped in the vicinity of attacked samples unlike the other models.
	Besides, only ARMA outputs a highly similar pattern to $\bm{Y}$: a non-attacked sample cluster in the center and attacked samples distributed around it.

	\subsection{Discussions \& Theoretical Comparisons} \label{sec:discuss}	
	As indicated earlier, two main approaches have been proposed for detecting and localizing the FDIAs: model-based and data driven approaches \cite{musleh2019survey}.
	Model-based approaches such as those in \cite{nudell2015real, khalaf2018joint, drayer2019cyber, luo2020interval, hasnat2020detection} do not require any historical datasets. Nevertheless, scalability, manual threshold optimization process, detection lags, model complexity, and localization resolution could hinder the usability of them for real applications.	
	For instance, results are not published in \cite{drayer2019cyber} and localization could only be done at the cluster level in \cite{nudell2015real}.
	Detection times are larger than a second in \cite{khalaf2018joint} and \cite{luo2020interval} for small test systems having 12 and 36 buses, respectively.	
	In their model-based detectors, authors of \cite{hasnat2020detection} utilize GSP techniques such as Local Smoothness (LS)  and Vertex-Frequency Energy Distribution (VFED). Nonetheless, they evaluate their method with an easily detectable attack by the classical LNRT based BDD methods which can conceal the actual performance of the the model.
	Specifically, they simulate the FDIAs using $z_a^i(t) = z_o^i(t) + (-1)^d . a . u . range$  where $d \sim \{0,1\}$ is a binary random variable (r.v.), $u \sim \U[0, 1]$ is an uniform r.v., $range = \max(z_o^i)-\min(z_o^i)$ and $a$ is scaler for the attack.
	For instance, if $z_o^i \sim \N(\mu_o,\sigma_o^2)$, expected values of the attacked data distribution become $\EX[\mu_a]=\mu_o$, and $\EX[\sigma_a]=6 a \sigma_o$ due to the product properties of uniform and normal distributions, where $\mu_o,\sigma_o$ and $\mu_a,\sigma_a$ tuples represent the mean and standard deviation of original and attacked data, respectively.
	The accuracy of localization for IEEE 118-bus test system when $a=4$, which makes $\EX[\sigma_a]=24\sigma_o$, are $85\%$ and $91\%$ for the LS and VFED techniques, respectively.
	These high accuries are not realistic since the scaler $a$ plays a significant role in simulation process.
	
	The data-driven methods, in contrast, present a better performance since historical datasets are growing and the  modeling capability of these algorithms is being increased \cite{musleh2019survey, sayghe2020survey}.
	For example, in their data-driven method in \cite{jevtic2018physics} researchers employ an LSTM model for each measurement in a 5-bus test system in which only one bus is under attack at a time to detect and localize the point-wise FDIAs.
	They report greater than $90\%$ accuracy for detection and localization of random, ramp, and scale attacks for low, medium, and high attack levels.
	However, the capability of this method for detection and localization of different FDIAs in large test cases has not been investigated.
	Besides, assigning an independent model to each measurement has two major drawbacks: (i) overall model complexity increases severely, and (ii) spatial correlations of the measurements are ignored totally.   
	
	In data-driven approaches, compatibility between the structure of collected data and architecture of the data-driven model is the primary factor on the performance of the model.
	For instance, DT, KNN, or MLP architectures could be better suited for a dataset having uncorrelated features from different spaces.
	Similarly, an RNN architecture might be more applicable to model the recurrent relations in a natural language data.
	A CNN architecture, in contrast, could be more favorable than GNN for an image data where pixel locality is well modeled in 2D Euclidean space.
	However, as demonstrated with Fig.~\ref{fig:euclid}, spatial correlations in power measurements data can only be captured in a non-Euclidean space dictated by the topology of the grid.
	For instance, if we had a hypothetical power grid like in Fig.~\ref{fig:euclid}, a CNN architecture could have comparable performances with ARMA.
	Nonetheless, for a power grid data collected from graph type structure, a GNN is more advantageous than other architectures as can be seen from the detection results in Table~\ref{tab:detection} and the localization distributions in Figs. \ref{fig:localization_swe} and \ref{fig:localization_nwe}.
	As for the GNN family, ARMA outperforms CHEB due to the fact that rational GFs implemented using the ARMA architecture provide more flexible frequency responses compared to the polynomial filters such as CHEB \cite{loukas2015distributed}.
	
	It is observed from our extensive experiments that the proposed ARMA based model performs better compared to other models for larger test cases.
	As an illustration, for the $95\%$ $F1$ threshold level, it outperforms the second best model CHEB by $5.64\%$, $8.56\%$, and $10.07\%$ in SW localization and by $8.78\%$, $11.87\%$, and $14.67\%$ in NW localization for IEEE 57-, 118-, and 300-bus systems, respectively.
	This difference is due to the fact that in larger and denser graphs, 
	(i) the spatial correlation between adjacent measurements becomes more dominant compared to the global correlations and 
	(ii) ARMA GFs better adapt to abrupt changes in the spectral domain compared to the polynomial ones.

\begin{figure}[h!]
		\centering
		\includegraphics[width=0.97\linewidth]{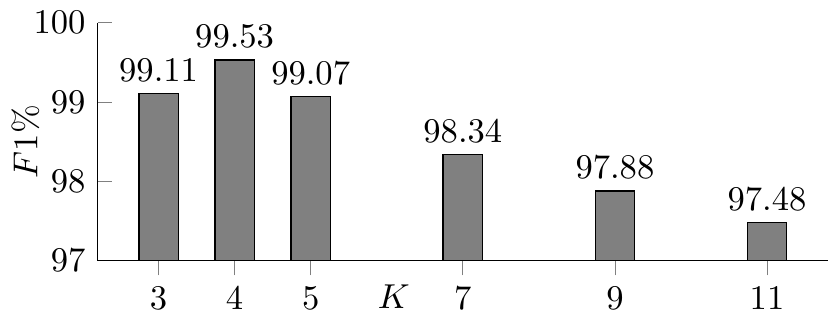}
		\caption{Detection performance vs. filter order of CHEB models for IEEE 300-bus system. Optimal results are obtained at $K=4$ as given in Table \ref{tab:hpo}.}
		\label{fig:order}
	\end{figure}

	As stated before, the output of each vertex $v$ only depends on its $K$-hop neighbors for a $K$-order polynomial GF.
	In other words, the output of $v$ is independent of the vertices beyond the $K$-hop neighbors for a $K$-order FIR GF \cite{shi2015infinite}.
	Thus, to capture the global characteristics of the graph, an FIR GF requires ``high'' order spectral response as can be seen from Fig.~\ref{fig:filters}.
	Nevertheless, due to the poor interpolation and extrapolation capabilities of the high order polynomials, it becomes sensitive to variations and may overfit to the training data \cite{bianchi2021graph}.
	To verify this characteristic, we fix the other parameters of CHEB GF at their optimal values tabulated in Table~\ref{tab:hpo} and train a CHEB model for the IEEE 300-bus test system for each $K \in \{5,7,9,11\}$.
	FDIA detection results in terms of $F1\%$ are depicted in Fig.~\ref{fig:order}. 
	Clearly, increasing $K$ beyond a certain point makes the model susceptible to variations such as noise, and thus, it can degrade the test set performance.	
	Note that similar conclusion can also be corroborated for the localization results.

	Bus level localization is a multi-label classification task and should be evaluated accordingly.
	Besides, performance metrics can cause inaccurate or misleading outcomes when they are not interpreted correctly.
	Namely, missing an attack (FN) could be much more severe than a false alarm (FP) when dealing with FDIAs due to their consequences.
	An example of localization results for a hypothetical model is given in Table~\ref{tab:ex_loc} with 4 samples in rows and 5 buses in columns where TP, FP, FN, and TN samples are highlighted with green, blue, red, and black colors, respectively.
	In addition, SW and NW localization results are given at the end of each row and columns in terms of accuracy $ACC = \frac{TP+TN}{TP + TN + FP + FN}$ and $F1$ percentages.
	\begin{table}[h!]
	\centering
	\caption{{SW} and {NW} localization example in $ACC\%$ and $F1\%$}
	\setlength{\tabcolsep}{3pt}
	\renewcommand{\arraystretch}{1.2}
	\newcolumntype{?}[1]{!{\vrule width #1}}
	\begin{tabular}{c ?{1pt} ccccc ?{1pt} c ?{1pt} c}
		& $n_1$   & $n_2$   & $n_3$    & $n_4$    & $n_5$   & $ACC_{sw}$ & $F1_{sw}$ \\
		\specialrule{1pt}{1pt}{1pt}
		$s_1$ & \tn{tn} & \fn{fn} & \tn{tn}  & \tn{tn}  & \tn{tn} & 80  & 0 \\
		$s_2$ & \tn{tn} & \fn{fn} & \tn{tn}  & \tn{tn}  & \tn{tn} & 80  & 0   \\
		$s_3$ & \tn{tn} & \fn{fn} & \fp{fp}  & \tn{tn}  & \tp{tp} & 60  & 50   \\
		$s_4$ & \tp{tp} & \fn{fn} & \tp{tp}  & \fp{fp}  & \tp{tp} & 60  & 75  \\
		\specialrule{1pt}{1pt}{1pt}
		\textit{$ACC_{nw}$} & 100   & 0       & 75       & 75       & 100     &     &    \\
		\specialrule{1pt}{1pt}{1pt}
		\textit{$F1_{nw}$}  & 100   & 0       & 66       & 0        & 100     &     & 
	\end{tabular}
	\label{tab:ex_loc}
	\end{table}
	The $ACC_{sw}$ is not a reliable metric since it can not properly take into account the distribution of errors. For instance, although $ACC_{sw}$ shows high accuracy for all the samples, it does not have any mechanism to mirror the faults at $n_2$ which can have
	serious consequences for the power system. 
	Comparing $F1$ with $ACC$ reveals that $F1$ has a better mechanism to evaluate the accuracy of the model.
	For instance, the $ACC_{sw} = 60\%$ for $s_3$ and $s_4$ since they have the same number of falsely predicted samples.
	$F1_{sw}$ metric, in contrast, yields $50\%$ for $s_3$ and $75\%$ for $s_4$ since $s_4$ includes 1 more TP compared to the $s_3$.
	Since our focus is to determine the localization of FDIAs, then $F1$ is the proper candidate to evaluate the accuracy of the model. The result and discussion reveals the supremacy of our model compared to DT, MLP, RNN, CNN, and CHEB models in terms of detection and localization of FDIAs.

	\section{Conclusion}\label{conclusion} 
	This work proposed a GNN based model by integrating the underlying graph topology of the grid and spatial correlations of its measurement data to jointly detect and localize the FDIAs in power systems while the full AC power flow equations are employed to address the physics of the network.
	Adopting IIR type ARMA GFs in its hidden layers, the proposed model is more flexible in frequency response compared to FIR type polynomial GFs, e.g., CHEB thanks to their rational type filter composition.
	Although our algorithm has better detection and localization performance compared to the state of the art CHEB model \cite{boyaci2021graph} in the literature, the improvement rate for localization is much higher than detection.
	Simulations performed on various standard test systems confirm that the performance of the proposed model in detecting FDIA exceeds the performance of CHEB model by 0.12\%, 0.68\%, and 0.38\% for IEEE 57-, 118-, and 300-bus, respectively.	
	The proposed model also outperforms the CHEB model in localizing the attacks (i.e., $95\%$ $F1$ threshold level) by $5.64\%$, $8.56\%$, and $10.07\%$ in SW localization and by $8.78\%$, $11.87\%$, and $14.67\%$ in NW localization for the same above-mentioned test systems, respectively.
	Furthermore, visualizing the intermediate layers for different approaches including those in literature corroborates the supremacy of the proposed model in detecting FDIA.

	\newcommand{\BIBentryALTinterwordspacing}{}
	\newcommand{\BIBentrySTDinterwordspacing}{}
	\bibliographystyle{IEEEtran}
	\bibliography{main}

	\vspace{-35pt}

	\begin{IEEEbiography}[{\includegraphics[width=1in,height=1.25in,clip,keepaspectratio]{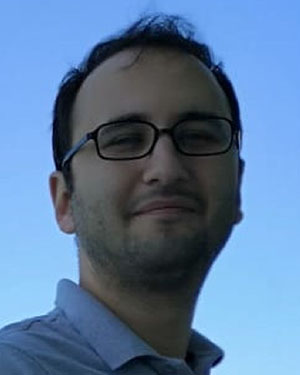}}]%
		{Osman Boyaci} (Student Member, IEEE) received the B.Sc. (Hons.) degree in electronics engineering in 2013 and in computer engineering (double major) in 2013 (Hons.) from Istanbul Technical University, Istanbul, Turkey. He received the M.Sc. degree in computer engineering at the same university in 2017. Currently, he is a Ph.D. candidate at Texas A\&M University working on graph neural network based cybersecurity in smart grids.
		
		His research interests include machine learning, artificial intelligence, and cybersecurity.
	\end{IEEEbiography}
	
	\vspace{-35pt}
	
	\begin{IEEEbiography}[{\includegraphics[width=1in,height=1.25in,clip,keepaspectratio]{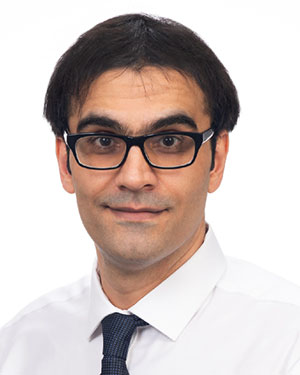}}]%
		{Mohammad Rasoul Narimani}
		(S'14-M'20) is an Assistant Professor in the College of Engineering at Arkansas State University. Before joining Arkansas State University, he was a postdoc at Texas A\&M University, College Station. He received the B.S. and M.S. degrees in electrical engineering from the Razi University and Shiraz University of Technology, respectively. He received the Ph.D. in electrical engineering from Missouri University of Science \& Technology. His research interests are in the application of optimization techniques to electric power systems.
	\end{IEEEbiography}
	
	\vspace{-40pt}
	
	\begin{IEEEbiography}[{\includegraphics[width=1in,height=1.25in,clip,keepaspectratio]{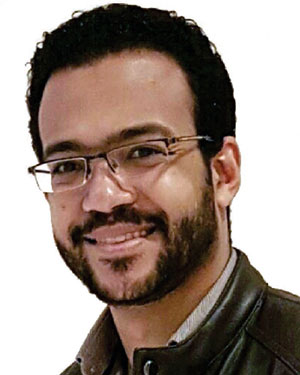}}]%
		{Muhammad Ismail} (S'10-M'13-SM'17) received the B.Sc. (Hons.) and M.Sc. degrees in Electrical Engineering (Electronics and Communications) from Ain Shams University, Cairo, Egypt, in 2007 and 2009, respectively, and the Ph.D. degree in Electrical and Computer Engineering from the University of Waterloo, Waterloo, ON, Canada, in 2013. He is currently an Assistant Professor with the Department of Computer Science, Tennessee Tech. University, Cookeville, TN, USA. He was a co-recipient of the Best Paper Awards in the IEEE ICC 2014, the IEEE Globecom 2014, the SGRE 2015, the Green 2016, the Best Conference Paper Award from the IEEE TCGCN at the IEEE ICC 2019, and IEEE IS 2020.  
	\end{IEEEbiography}
	
	\vspace{-40pt}
	
	\begin{IEEEbiography}[{\includegraphics[width=1in,height=1.25in,clip,keepaspectratio]{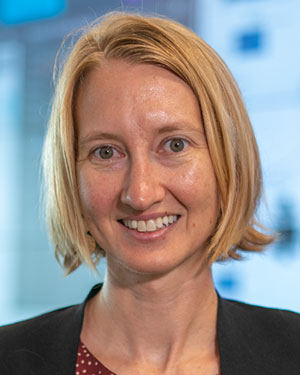}}]%
		{Katherine R. Davis} (S'05-M'12-SM'18) received the B.S. degree from the University of Texas atAustin, Austin, TX, USA, in 2007, and the M.S. and Ph.D. degrees from the University of Illinois at Urbana-Champaign, Champaign, IL, USA, in 2009 and 2011, respectively, all in electrical engineering. She is currently anAssistant Professor of Electrical and Computer Engineering at TAMU.
	\end{IEEEbiography}
	
	\vspace{-20pt}
	
	\begin{IEEEbiography}[{\includegraphics[width=1in,height=1.25in,clip,keepaspectratio]{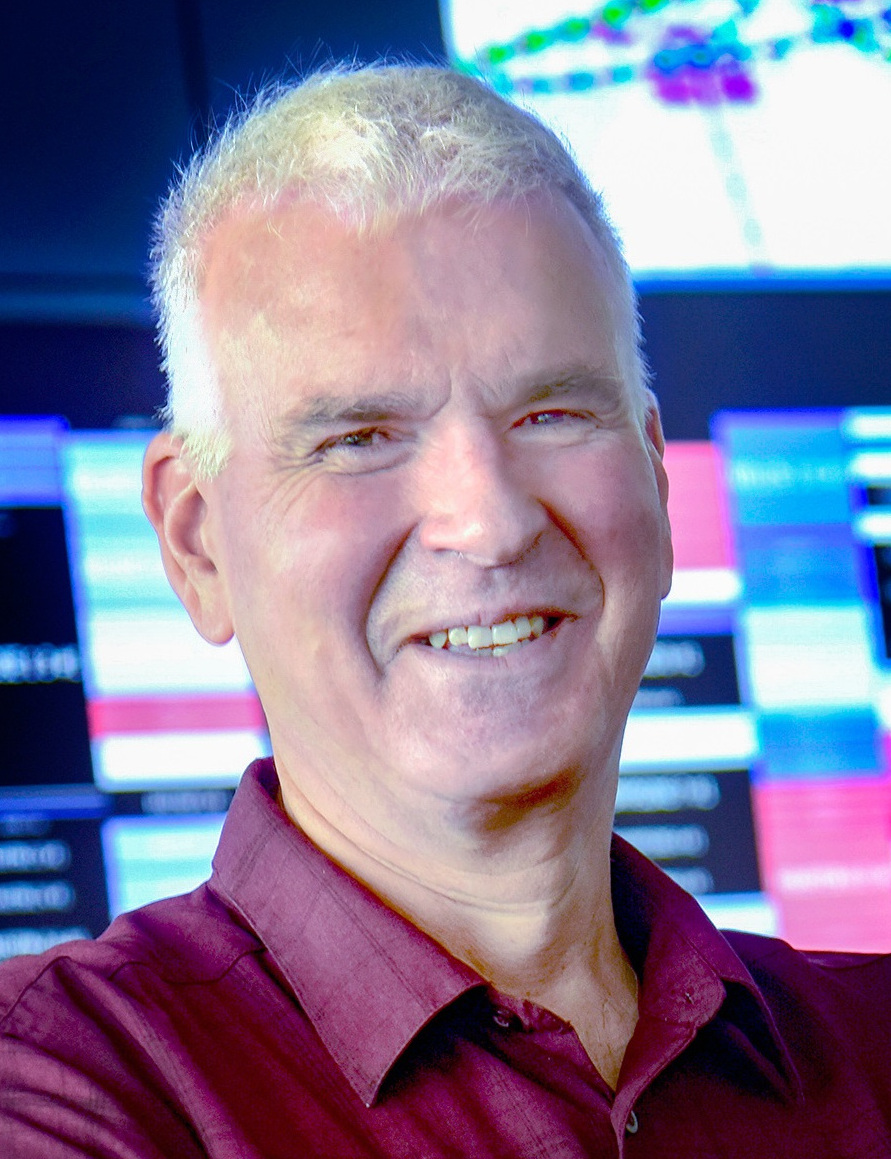}}]%
		{Dr. Thomas J. Overbye} (S'87-M'92-SM'96-F'05) received B.S., M.S., and Ph.D. degrees in electrical engineering from the University of Wisconsin Madison, Madison, WI, USA. He is currently with Texas A\&M University where he is a Professor and holder of the  O'Donnell Foundation Chair III. 
	\end{IEEEbiography}
	
	\vspace{-20pt}
	
	\begin{IEEEbiography}[{\includegraphics[width=1in,height=1.25in,clip,keepaspectratio]{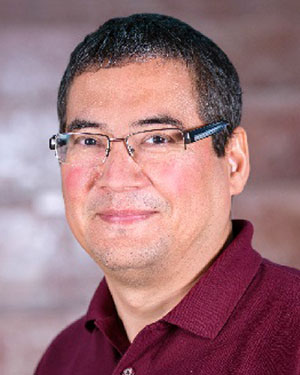}}]%
		{Dr. Erchin Serpedin} is a professor in the Electrical and Computer Engineering Department at Texas A\&M University in College Station. Dr. Serpedin is the author of four research monographs, one textbook, 17 book chapters, 170 journal papers, and 270 conference papers. His current research interests include signal processing, machine learning, artificial intelligence, cyber security, smart grids, and wireless communications. He served as an associate editor for more than 12 journals, including journals such as the IEEE Transactions on Information Theory, IEEE Transactions on Signal Processing, IEEE Transactions on Communications, IEEE Signal Processing Letters, IEEE Communications Letters, IEEE Transactions on Wireless Communications, IEEE Signal Processing Magazine, and Signal Processing (Elsevier), and as a Technical Chair for six major conferences. He is an IEEE Fellow. 
	\end{IEEEbiography}

\end{document}